\documentclass{article}

\usepackage[noend]{algorithmic}

\usepackage{balance}

\usepackage{amsmath,amsfonts,bm}
\usepackage{mathtools}
\usepackage{soul}

\newcommand{\ignore}[1]{}

\newcommand{\captionn}[1]{{\em (#1)}}

\DeclarePairedDelimiter\abs{\lvert}{\rvert}%
\DeclarePairedDelimiter\norm{\lVert}{\rVert}%

\makeatletter
\let\oldabs\abs
\def\abs{\@ifstar{\oldabs}{\oldabs*}}
\let\oldnorm\norm
\def\norm{\@ifstar{\oldnorm}{\oldnorm*}}
\makeatother

\newcommand{\figlabel}[1]{\label{Fi:#1}}
\newcommand{\tablabel}[1]{\label{Ta:#1}}

\def\Figref#1{Figure~\ref{Fi:#1}}

\def\Tabref#1{Table~\ref{Ta:#1}}

\def\Secref#1{Section~\ref{#1}}

\def\Appref#1{Appendix~\ref{#1}}

\def\eqref#1{equation~\ref{#1}}
\def\Eqref#1{Equation~\ref{#1}}

\def\Algref#1{Algorithm~\ref{#1}}

\def\floor#1{\lfloor #1 \rfloor}
\def\1{\bm{1}}

\def\rmG{{\mathbf{G}}}

\def\va{{\bm{a}}}

\def\vp{{\bm{p}}}

\def\vs{{\bm{s}}}

\def\vy{{\bm{y}}}

\DeclareMathAlphabet{\mathsfit}{\encodingdefault}{\sfdefault}{m}{sl}
\SetMathAlphabet{\mathsfit}{bold}{\encodingdefault}{\sfdefault}{bx}{n}

\def\sN{{\mathbb{N}}}

\def\sR{{\mathbb{R}}}

\def\sW{{\mathbb{W}}}
\def\sX{{\mathbb{X}}}
\def\sY{{\mathbb{Y}}}

\newcommand{\E}{\mathbb{E}}

\newcommand{\R}{\mathbb{R}}

\newcommand{\normlone}{L^1}

\DeclareMathOperator*{\argmin}{arg\,min}

\newcommand{\scode}[1]{\ensuremath{\mathtt{#1}}}

\newcommand{\dataset}{\mathcal{D}}

\newcommand{\lstm}{\ensuremath{\scode{LSTM}}}
\newcommand{\deeptyper}{\ensuremath{\scode{DeepTyper}}}
\newcommand{\gcn}{\ensuremath{\scode{GCN}}}
\newcommand{\ggnn}{\ensuremath{\scode{GGNN}}}
\newcommand{\gnt}{\ensuremath{\scode{GNT}}}

\newcommand{\mods}{\ensuremath{\bm{\delta}}}

\usepackage[mathscr]{euscript}

\usepackage{booktabs}   
\newcommand{\tra}[1]{\renewcommand{\arraystretch}{#1}}

\usepackage{subcaption} 

\usepackage{pifont} 

\usepackage{multirow}

\usepackage{tikz}
\usetikzlibrary{positioning}
\usetikzlibrary{calc}
\usetikzlibrary{decorations.pathreplacing}
\usetikzlibrary{arrows.meta}
\usetikzlibrary{plotmarks}
\usetikzlibrary{snakes,automata,backgrounds,petri}
\usepackage{tikz-qtree}
\usetikzlibrary{matrix}
\usetikzlibrary{arrows,shapes,patterns}
\tikzset{
    >=stealth',
    boxx/.style={
           rectangle,
           draw=black,
           text width=6.5em,
           minimum height=2em,
           text centered},
}
\usepackage{forest}

\usepackage{listings}

\lstdefinelanguage{JavaScript}{
  basicstyle=\footnotesize\ttfamily,
  keywords={typeof, new, true, false, catch, function, return, null, catch, switch, var, if, in, while, do, else, case, break},
  keywordstyle=\color{blue}\bfseries,
  ndkeywords={class, export, boolean, throw, implements, import, this},
  ndkeywordstyle=\color{darkgray}\bfseries,
  identifierstyle=\color{black},
  sensitive=false,
  comment=[l]{//},
  morecomment=[s]{/*}{*/},
  commentstyle=\color{purple}\ttfamily,
  stringstyle=\color{red}\ttfamily,
  morestring=[b]',
  morestring=[b]"
}

\usepackage{amsthm}
\theoremstyle{definition}
\newtheorem{definition}{Definition}[section]

\usepackage{breakurl}
\usepackage[breaklinks]{hyperref}

\usepackage[accepted]{icml2020}

\icmltitlerunning{Adversarial Robustness for Code}

\begin{document}

\definecolor{darkgreen}{rgb}{0.0, 0.5, 0.0}
\definecolor{darkblue}{rgb}{0.0, 0.0, 0.5}
\definecolor{myorange}{RGB}{255, 193, 7}

\DeclareRobustCommand\graybox{\tikz{\draw[fill=gray!20!white] (6.69,0) rectangle ++(0.2,0.2);}}

\DeclareRobustCommand\bluebox{\tikz{\draw[fill=blue!30!white] (6.69,0) rectangle ++(0.2,0.2);}}

\DeclareRobustCommand\orangebox{\tikz{\draw[fill=myorange!80!white] (6.69,0) rectangle ++(0.2,0.2);}}

\DeclareRobustCommand\whitebox{\tikz{\draw[fill=white] (6.69,0) rectangle ++(0.2,0.2);}}

\twocolumn[
\icmltitle{Adversarial Robustness for Code}



\icmlsetsymbol{equal}{*}

\begin{icmlauthorlist}
\icmlauthor{Pavol Bielik}{eth}
\icmlauthor{Martin Vechev}{eth}
\end{icmlauthorlist}

\icmlaffiliation{eth}{Department of Computer Science, ETH Z\"{u}rich, Switzerland}

\icmlcorrespondingauthor{Pavol Bielik}{pavol.bielik@inf.ethz.ch}

\icmlkeywords{Robust Code, Adversarial Robustness, Adversarial Training, Source Code, Programs, Learning to Abstain, Representation Refinement}

\vskip 0.3in
]



\printAffiliationsAndNotice{}  

\begin{abstract}
Machine learning and deep learning in particular has been recently used to successfully address many tasks in the domain of code such as finding and fixing bugs, code completion, decompilation, type inference and many others. However, the issue of adversarial robustness of models for code has gone largely unnoticed. In this work, we explore this issue by: \captionn{i} instantiating adversarial attacks for code (a domain with discrete and highly structured inputs), \captionn{ii} showing that, similar to other domains, neural models for code are vulnerable to adversarial attacks, and \captionn{iii} combining existing and novel techniques to improve robustness while preserving high accuracy.
\end{abstract}

\section{Introduction}
Recent years have seen an increased interest in using deep learning to train models of code for a wide range of tasks including code completion~\cite{brockschmidt2018generative, Li:2018}, code captioning~\citep{Alon:2019, allamanis2016convolutional, fernandes2018structured}, code classification~\cite{Mou:2016, Zhang:2019} and bug detection~\cite{allamanis2018learning, DeepBugs, Li:2019}. 
Despite substantial progress on training accurate models of code, the issue of robustness has been overlooked.
Yet, this is a very important problem shown to affect neural models in different domains~\cite{GoodfellowSS14, Szegedy:2014, PapernotMSH16}.

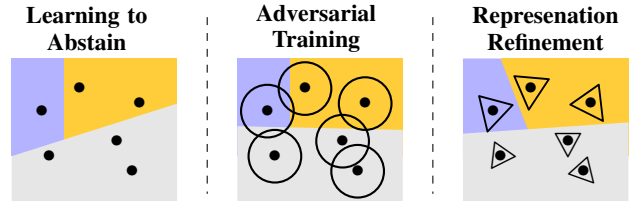
\begin{figure}[t]
\centering
\def\radius{0.35}
\begin{tikzpicture}[]

\node[font=\footnotesize, text width=2cm, align=center] (abstain) at (0,0) {\bf Learning to \\ \bf Abstain};

\begin{scope}
\fill[myorange!80!white] (-0.3, -0.4) -- (-0.3, -1.7) -- (1.2, -1.7) -- (1.2, -0.4);
\fill[blue!30!white] (-1,-1.7) -- (-1, -0.4) -- (-0.3, -0.4) -- (-0.3, -1.7);
\fill[gray!20!white] (-1,-1.7) -- (1.2,-1.0) -- (1.2, -2.3) -- (-1, -2.3) -- (-1, -1.5);

\node (a) at (-0.1, -0.8)  {\textbullet};
\node (b) at (-0.6, -1.1)  {\textbullet};
\node (c) at (0.7, -1.0)  {\textbullet};

\node (d) at (0.4, -1.5)  {\textbullet};
\node (e) at (-0.5, -1.7)  {\textbullet};
\node (f) at (0.6, -1.9)  {\textbullet};
\end{scope}

\node[font=\footnotesize, text width=2cm, align=center, right=of abstain, xshift=-0.2cm] (adversarial) {\bf Adversarial \\ \bf Training};

\begin{scope}[xshift=3cm]
\fill[myorange!80!white] (-0.3, -0.4) -- (-0.3, -1.7) -- (1.2, -1.7) -- (1.2, -0.4);
\fill[blue!30!white] (-1,-1.7) -- (-1, -0.4) -- (-0.3, -0.4) -- (-0.3, -1.7);
\fill[gray!20!white] (-1,-1.3) -- (1.2,-1.35) -- (1.2, -2.3) -- (-1, -2.3) -- (-1, -1.5);

\node (a) at (-0.1, -0.8)  {\textbullet};
\node (b) at (-0.6, -1.1)  {\textbullet};
\node (c) at (0.7, -1.0)  {\textbullet};

\node (d) at (0.4, -1.5)  {\textbullet};
\node (e) at (-0.5, -1.7)  {\textbullet};
\node (f) at (0.6, -1.9)  {\textbullet};

\draw[thick] (a) circle (\radius);
\draw[thick] (b) circle (\radius);
\draw[thick] (c) circle (\radius);
\draw[thick] (d) circle (\radius);
\draw[thick] (e) circle (\radius);
\draw[thick] (f) circle (\radius);
\end{scope}

\node[font=\footnotesize, text width=2cm, align=center, right=of adversarial, xshift=-0.2cm] (refine) {\bf Represenation \\ \bf Refinement};

\begin{scope}[xshift=6cm]
\fill[myorange!80!white] (-0.5, -0.4) -- (-0.5, -1.7) -- (1.2, -1.7) -- (1.2, -0.4);
\fill[blue!30!white] (-1,-1.7) -- (-1, -0.4) -- (-0.5, -0.4) -- (0, -1.7);
\fill[gray!20!white] (-1,-1.4) -- (1.2,-1.25) -- (1.2, -2.3) -- (-1, -2.3) -- (-1, -1.5);

\node (a) at (-0.1, -0.8)  {\textbullet};
\node (b) at (-0.6, -1.1)  {\textbullet};
\node (c) at (0.7, -1.0)  {\textbullet};

\node (d) at (0.4, -1.5)  {\textbullet};
\node (e) at (-0.5, -1.7)  {\textbullet};
\node (f) at (0.6, -1.9)  {\textbullet};

\node[scale=1.8, rotate=-5] at (a)  {$\bigtriangledown$};
\node[scale=1.8, rotate=-15] at (b)  {$\bigtriangledown$};
\node[scale=1.8, rotate=35] at (c)  {$\bigtriangledown$};

\node[scale=1.3, rotate=0] at (d)  {$\bigtriangledown$};
\node[scale=1.3, rotate=-35] at (e)  {$\bigtriangledown$};
\node[scale=1.3, rotate=42] at (f)  {$\bigtriangledown$};

\end{scope}

\draw [dashed] (1.6,0.15) -- (1.6,-2.2);
\draw [dashed] (4.6,0.15) -- (4.6,-2.2);

\end{tikzpicture}
\vspace{-2.3em}
\caption{Illustration of the three key components used in our work. Each point represents a sample, \graybox{} is a region where model abstains from making predictions, \bluebox{} and \orangebox{} are regions of model prediction, $\bm{\bigcirc}$ is the space of valid modifications for a~given sample, and $\bm{\rhd}$ is the learned (reduced) space of valid modifications.}
\figlabel{intro}
\vspace{-0.15in}
\end{figure}

\paragraph{Challenges in modeling code}
In our work, we focus on tasks that compute program properties (e.g., type inference), usually addressed via handcrafted static analysis, but for which a number of recent neural models with high accuracy have been introduced~\cite{DeepTyper:18, Schrouff:2019, NL2Type:19}.
Unsurprisingly, as these works do not consider adversarial robustness,  their adversarial accuracy can drop significantly.
However, training both \emph{robust and accurate} models of code in this setting is non-trivial and requires one to address several key challenges: \captionn{i} programs are highly structured and long, containing hundreds of lines of code, \captionn{ii} a~single discrete program change can affect the prediction of a large number of properties and is much more disruptive than a~slight continuous perturbation of a pixel value, and \captionn{iii} the property prediction problem is usually undecidable (hence, static analyzers approximate the ideal solution).

\paragraph{Accurate and robust models of code}
As a first step to address these challenges, we propose a novel method that combines three key components, illustrated in \Figref{intro} -- as we show, all of these contribute to achieving accurate and robust models of code.
First, we train a model that \emph{abstains}~\cite{DeepGamblers:19} from making a prediction when uncertain, effectively partitioning the dataset into two parts: one part where the model makes predictions (\bluebox{}, \orangebox{}) that should be \emph{accurate} and \emph{robust}, and one~(\graybox{}) where the model abstains and it is enough to be \emph{robust}. Second, we instantiate \emph{adversarial training}~\cite{GoodfellowSS14} to the domain of code. 
Third, we develop a~new method~to \emph{refine the representation} used as input to the model by learning the parts of the program relevant for the~prediction.
This reduces the number of places that affect the prediction and helps to make adversarial training for code effective. 
Finally, we create a new algorithm that trains multiple models, each learning a~specialized representation that makes robust predictions on a different subset of the dataset.

We instantiate our approach to the type prediction task and show its effectiveness -- we train a model that improves robustness by $15\%$ while preserving high accuracy.

\section{Accurate and Robust Models of Code}\label{overview}

\tikzset{round left paren/.style={ncbar=0.5cm,out=120,in=-120}}
\tikzset{round right paren/.style={ncbar=0.5cm,out=60,in=-60}}

\begin{figure*}[t]
\centering

\begin{tikzpicture}[scale=0.88, every node/.append style={transform shape},
 trim left=-2.35cm,
 arrow/.style = {thick,-stealth},
 ncbar angle/.initial=90,
 ncbar/.style={
        to path=(\tikztostart)
        -- ($(\tikztostart)!#1!\pgfkeysvalueof{/tikz/ncbar angle}:(\tikztotarget)$)
        -- ($(\tikztotarget)!($(\tikztostart)!#1!\pgfkeysvalueof{/tikz/ncbar angle}:(\tikztotarget)$)!\pgfkeysvalueof{/tikz/ncbar angle}:(\tikztostart)$)
        -- (\tikztotarget)
    },
 ncbar/.default=0.5cm,
]

\node (base) at (0, 0) { 
\begin{lstlisting}[language=JavaScript, mathescape]
(hex$^\scode{str}$, radix$^\scode{num}$) => {
  v$^\scode{num}$ = parseInt$^\scode{num}$(
    hex$^\scode{str}$.substring$^\scode{str}$(1$^\scode{num}$), 
    radix$^\scode{num}$
  );
  red$^\scode{num}$ = v$^\scode{num}$ >>$^\scode{num}$ 16$^\scode{num}$;
  ...
\end{lstlisting}
};

\node[above=of base, yshift=-1.1cm, text width=5.5cm, align=center] (reject_label) { \footnotesize \captionn{a} \bf Training Dataset $\dataset\!=\!\{ (x_j, y_j) \}_{j=1}^N$};



%


\node[right=of base, xshift=-0.3cm] (reject) { 
\begin{lstlisting}[language=JavaScript, mathescape]
(hex, radix) => {
  v$^{\color{darkgreen}\scode{num, 1.0}}$ = parseInt$^{\color{darkgreen}\scode{num, 1.0}}$(
      hex.substring$^{\color{darkgreen}\scode{str, 0.9}}$(1$^{\color{darkgreen}\scode{num, 1.0}}$), 
      radix
    );
  red = v >> 16$^{\color{darkgreen}\scode{num, 1.0}}$;
$~$
\end{lstlisting}
};


\node[below=of reject, yshift=1.34cm, font=\footnotesize, text width=7.5cm, align=center] { Learn selection function $g_h$ that makes a prediction \\  only if confident enough and abstains otherwise.};

\node[above=of reject, yshift=-1.1cm, text width=6cm, align=center] (reject_label) { \footnotesize \captionn{b} \bf Learning to Abstain  (Appendix B)};

\node[right=of reject, xshift=-0.2cm] (adversarial) { 
\begin{lstlisting}[language=JavaScript, mathescape]
(color$^{\color{red}{\scode{num, 0.9}}}$, radix) => {
  v$^{\scode{num, 0.4}}$ = parseInt$^{\scode{num, 0.6}}$(
      color.substring$^{\color{red}{\scode{bool, 0.9}}}$(1$^{\color{darkgreen}\scode{num, 1.0}}$), 
      radix
    );
  red = v >> 16$^{\color{darkgreen}\scode{num, 1.0}}$;
  $\footnotesize \qquad \quad \mods = [\scode{rename~hex} \rightarrow \scode{color}]$
\end{lstlisting}
};


\node[below=of adversarial, yshift=1.0cm, font=\footnotesize] { Train with the worst case modifications $x + \mods$};

\node[above=of adversarial, yshift=-1.1cm, text width=6cm, align=center] (adversarial_label) { \footnotesize \captionn{c} \bf Adversarial Training (Appendix C)};

\draw [arrow, shorten >= 0.0cm, shorten <= 0.0cm] (reject) --node [above, align=center, yshift=-0.0cm, font=\footnotesize] { $\langle f, g_h \rangle$ } (adversarial);


\draw [arrow] ($(adversarial.south)+(0.8,-0.6)$) --node [left, align=center, font=\footnotesize] { $\langle f, g_h \rangle$ } ($(adversarial.south)+(0.3,-1.4)$);

\draw [arrow] ($(adversarial.south)+(0.5,-1.4)$) --node [right, align=center, font=\footnotesize, text width=2cm] { Retrain with \\ abstraction $\alpha$ } ($(adversarial.south)+(1.0,-0.6)$);



\node[below=of reject, xshift=3.1cm, yshift=0.4cm, text width=6cm, align=center] (refinement_label) { \footnotesize \captionn{d} \bf Representation Refinement (\Secref{refinement})};

\node[below=of refinement_label, yshift=-1.5cm] (refinement_note) { \footnotesize Learned abstraction $\alpha$ used to predict the \scode{parseInt} return type};

\node[below=of refinement_label, xshift=-1.0cm, yshift=1.1cm] (alpha1) { 
\begin{lstlisting}[language=JavaScript, mathescape]
(num, radix) => {
  v = parseInt(
      num.substring(1), 
      radix
    );
  red = v >> 16;
\end{lstlisting}
};

\node[right=of alpha1, xshift=-0.5cm, yshift=0.0cm] { 
\begin{lstlisting}[language=JavaScript, mathescape]
parseInt(
  _, 
  _
);
\end{lstlisting}
};

\node[left=of alpha1, xshift=0.85cm] (alpha_label) {$\alpha$};
\draw [thin] ($(alpha_label.east)+(0.3,-1.2)$) to [round left paren ] ($(alpha_label.east)+(0.3,1.2)$);
\draw [thin] ($(alpha_label.east)+(4.4,-1.2)$) to [round right paren ] ($(alpha_label.east)+(4.4,1.2)$);
\node[right=of alpha1, xshift=-1.0cm] {$=$};

\node[left=of alpha1, xshift=-2.0cm, yshift=0.0cm] (apply) { 
\begin{lstlisting}[language=JavaScript, mathescape]
(hex, radix) => {
  v = parseInt$_{\color{darkblue}\scode{<num>}}$(
     hex.substring(1$_{\color{darkblue}\scode{<num>}}$), 
     radix
   );
  red = v >> 16$_{\color{darkblue}\scode{<num>}}$;
\end{lstlisting}
};

\node[above=of apply, xshift=0.0cm, yshift=-1.1cm, text width=7cm, align=center] (refinement_label) { \footnotesize \captionn{e} \bf Apply \& Train Next Model (\Secref{training})};

\node[left=of refinement_note, xshift=-0.8cm] { \footnotesize Annotate programs in $\dataset$ with predicted types};

\draw [arrow, shorten >= 0.2cm, shorten <= 0.2cm, transform canvas={xshift=0.2cm}] ($(apply.east)+(1.5,0)$) --node [above, align=center, yshift=-0.55cm, text width=2.4cm] {\footnotesize  \scode{robust} \scode{model} \\[0.4em] $m_i = \langle f, g_h, \alpha \rangle$} (apply.east);

\draw [arrow, shorten >= 0.0cm, shorten <= 0.0cm] (base) --node [above, align=center, yshift=0.0cm] {\footnotesize  $\dataset$} (reject);

\draw [arrow, shorten >= 0.0cm, shorten <= 0.0cm] (1.8,-1.9) --node [above, align=center, xshift=-0.7cm, yshift=-0.1cm] {\footnotesize  $\dataset_{i+1} \subset \dataset_{i}$} (2.4,-1.4);

\end{tikzpicture}
\vspace{-0.6cm}
\caption{Overview of the main steps of our approach for learning accurate and adversarially robust models of code.}
\figlabel{overview}
\end{figure*}
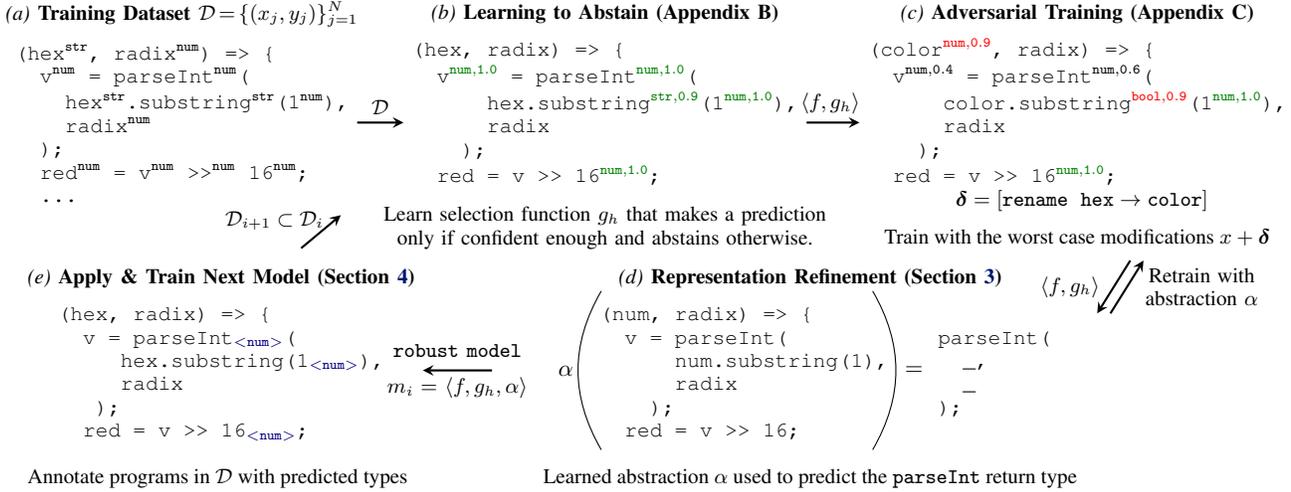

In this section, we present an overview of our approach.
Without loss of generality, we define an input program $p$ to be a sequence of words $p = w_{1:n}$. The words can correspond to a tokenized version of the program, nodes in an abstract syntax tree corresponding to~$p$ or other suitable program representations. Further, let $l\in\sN$ be a~position in the program $p$ that corresponds to a word $w_l\!\in\!\sW$.
A~training dataset $\dataset\!=\!\{ (x_j, y_j) \}_{j=1}^N$ contains~a~set of samples, where $x \in \sX$ is an input tuple $x =\langle p, l \rangle$ consisting of a~program $p$ and a~position in the program $l$, while $y\!\in\!\sY$ contains the ground-truth label.
As an example, the code snippet in \Figref{overview}\textit{a} contains 12 different samples $(x, y)$, one for each position where a prediction should be made (annotated with their ground-truth types $y$).

Our goal is to learn a function $f\colon \sX\!\rightarrow\!\sR^{\abs{\sY}}$, represented as a neural network, which for a given input program and a position in the program, computes the probability distribution over the labels.
The model's prediction then corresponds to the label with the highest probability according to $f$.

\vspace{-0.3em}
\paragraph{Step 1: Augment the model with an (un)certainty score}
We start by augmenting the standard neural model $f$ with an option to abstain from making a~prediction.
To achieve this, we adopt the recently proposed approach by \cite{DeepGamblers:19} and introduce a~selection function $g_h\colon \sX\!\rightarrow\!\sR$, which measures model certainty.
Then, the model is defined to make a prediction only if $g_h$ is confident enough ($g_h(x)~\geq~h$) and abstain from making a prediction otherwise.
Here, $h \in \sR$ is an associated threshold that controls the desired level of confidence.
For example, using a high threshold $h~=~0.9$, the model learns to make only five predictions for the program in \Figref{overview}\textit{b} and will abstain from uncertain predictions such as  predicting parameter types.

The first insight from our work is that allowing the model to abstain is beneficial for achieving robustness.
This step leads to simpler models, since learning to abstain is easier than learning to predict the correct label. This is in contrast with forcing the model to learn the correct label for \emph{all} samples, which is infeasible for most practical~tasks.

\vspace{-0.3em}
\paragraph{Step 2: Adversarial training}
Next, we instantiate adversarial training to the domain of code.
Concretely, let $\Delta(x)$ be a set of valid modifications of sample $x$ and let $x + \mods$ denote a~new input obtained by applying the modifications in $\mods \subseteq \Delta(x)$ to $x$.
As a~concrete example, \Figref{overview}\textit{c} shows a refactoring of the program from \Figref{overview}\textit{b} by renaming \scode{hex} to \scode{color}. Even though this change does not affect the types in the program, the model suddenly predicts incorrect types for both the \scode{color} parameter and the \scode{substring} function. Further, even though the types of \scode{parseInt} and \scode{v} are still correct, the model became much more uncertain.

Intuitively, our goal is to address this issue and to ensure that the model is robust for all valid modifications $\mods\!\subseteq\!\Delta(x)$ -- when evaluated on $x + \mods$, the model either abstains or predicts the correct label.
Concretely, we use adversarial training~\cite{GoodfellowSS14}, which instead of minimizing the expected loss on the original distribution $\E_{(x, y) \sim D}[ \ell((f, g_h)(x), y)]$ as usually done in standard training, minimizes the expected \textit{adversarial loss}:
\begin{equation}\label{adversarial_expected_loss}
\E_{(x, y) \sim D}[ \max_{\mods \subseteq \Delta(x)} \ell((f, g_h)(x + \mods), y)]
\end{equation}
That is, we minimize the worst case loss obtained by applying a valid modification to the original sample $x$.
Similar to other domains, the main challenge in this setting is solving the inner $\max_{\mods \subseteq \Delta(x)}$ efficiently for the domain of code.

\paragraph{Standard adversarial training is insufficient}
Although adversarial training has been successfully applied in many domains~\cite{madry2018towards, wong18a, sinha2018certifiable, raghunathan2018certified}, in our work we show that for code, adversarial training alone is insufficient to achieve model robustness.
The key reason is that, existing neural models of code typically process \textit{the entire program} which can contain hundreds of lines of code.
This is problematic as it means that any program change will affect \textit{all} predictions and there can be infinitely many program changes in $\Delta(x)$.
Further, a~single discrete program change is much more disruptive in affecting the model than a slight continuous perturbation of a pixel value.
At the same time, while not sufficient, in our evaluation we show that adversarial training can be used to improve robustness by $0$ to $7\%$, depending on the model architecture.

\paragraph{Step 3: Representation refinement}
To address the issue that adversarial training alone does not work well, we develop a novel technique that: (i) learns which parts of the input program are relevant for the given prediction, and (ii) refines the model representation such that only relevant program parts are used as input to the neural network. Essentially, the technique automatically learns an abstraction $\alpha$ which given a program, produces a relevant representation of that program. \Figref{overview}\textit{d} shows an example of a possible abstraction $\alpha$ that takes as input the entire program but keeps only parts relevant for predicting the type of \scode{parseInt} -- it is a method call with name \scode{parseInt} which has two arguments.
To learn the abstraction~$\alpha$, we first represent programs as graphs and then phrase the refinement task as an optimization problem that minimizes the number of graph edges, while ensuring that the accuracy of the model before and after applying $\alpha$ stays roughly the same.

Finally, we apply adversarial training, but this time on the abstraction $\alpha$ obtained via representation refinement, resulting in new functions $f$ and $g_h$. Overall, this results in an adversarially robust model $m_i = \langle f, g_h, \alpha \rangle$.

\paragraph{Step 4: Learning accurate models}
Although the model~$m_i$ is robust, it provides predictions only for a subset of the samples for which it has enough confidence (i.e., $g_h(x)\!\geq\!h$).
To increase the ratio of samples for which our approach makes a prediction (i.e., does not abstain), we perform two steps: \captionn{i} generate a new dataset $\dataset_{i+1}$ by annotating the program with the predictions made by the learned model $m_i$, and removing successfully predicted samples, and \captionn{ii} learn another model $m_{i+1}$ on the new dataset $\dataset_{i+1}$.
We repeat this process for as long as the new learned model predicts some of the samples in $\dataset_{i+1}$.

Training multiple models is beneficial because: \captionn{i} the models are easier to train as well as easier to make robust as they do not try to learn all predictions, \captionn{ii} it allows conditioning on the predictions learned by earlier models which helps both interpretability and robustness.
For example, the model $m_{i+1}$ can learn that the left hand side of the assignment \scode{v\!=\!parseInt} has the same type as the right hand side, since the type of \scode{parseInt} was already predicted by $m_i$.
Interestingly, if we think of each model as a learned set of rules, we can essentially apply the models to a given program in a~fixed point style (similar to how a traditional sound static analysis works), and 
\captionn{iii} each model learns a different representation $\alpha$ that is specialized for the predictions it makes. For example, while predicting the type of \scode{parseInt} is independent of the argument values ({\footnotesize \scode{parseInt(\_,~\_)}}), predicting the second argument type is not ({\footnotesize \scode{parseInt(\_,~radix)}}).
Using a~single abstraction to predict both would lead to either reduced robustness or accuracy, depending on which abstraction is~used.

\paragraph{Summary}
Given a training dataset $\dataset$, our approach learns a~set of robust models, each of which makes robust predictions for a different subset of $\dataset$.
To achieve this, we extend existing neural models of code with three key components -- the ability to abstain (with associated uncertainty score), adversarial training, and learning to refine the representation.
Given the limited space, we provide formal description of the the first two components that learn to abstain and apply adversarial training for code in Appendix B and Appendix C, respectively.
Next, we formally describe the novel components -- learning to refine the representation (\Secref{refinement}) and present our training algorithm that combines all of them together (\Secref{training}).

\section{Learning to Refine Representations}\label{refinement}
As motivated in \Secref{overview}, a key issue with many existing neural models for code is that the model prediction $f(x)$ depends on the \textit{full} program $p$, even though only small parts of $p$ are typically relevant. We address this issue by learning an abstraction $\alpha$ that takes as input $p$ and produces only the parts relevant for the prediction. That is, $\alpha$ refines the representation given as input to the neural model.

\paragraph{Overview}
Our method works as follows: \captionn{i}~we convert the program into a graph representation, \captionn{ii} then define the model to be a graph neural network (e.g., \cite{velickovic2018graph, kipf2017semi, SGC, li2016gated}), which at a high level works by propagating and aggregating messages along graph edges, \captionn{iii} because dependencies in graph neural networks are defined by the structure of the graph (i.e., the edges it contains), we phrase the problem of refining the representation as an optimization problem which removes the maximum number of graph edges (i.e., removes the maximum number of dependencies) without degrading model accuracy, and \captionn{iv}  we show how to solve the optimization problem efficiently by transforming it to an integer linear program~(ILP).

\paragraph{From programs to graphs}
Following prior works, we represent programs using their corresponding abstract syntax trees (AST).
These are further transformed into graphs, as done in~\cite{allamanis2018learning, brockschmidt2018generative}, by including additional edges.

\begin{definition}{(Directed Graph)}
A directed graph is a tuple $G = \langle V, E, \xi_V, \xi_E \rangle$ where $V$ denotes a set of nodes, $E \subseteq V^2$ denotes a set of directed edges, $\xi_V\colon V \rightarrow~\sN^k$ is a mapping from nodes to their associated attributes and $\xi_E\colon E \rightarrow \sN^m$ is a mapping from edges to their attributes.
\end{definition}

We associate two attributes with each node~--~\emph{type} which corresponds to the type of the AST node (e.g., \scode{Block}, \scode{Identifier}, \scode{BinaryExpression}, etc.) and \emph{value} associated with the AST node (e.g., \scode{+, -, 0, 1, "GET",} \scode{x, data}, etc.).
For edges we use a~single attribute the edge \textit{type}, which can be:
\captionn{i} \emph{ast}, for the edges that correspond to those included in the AST, \captionn{ii} \emph{last usage}, for edges introduced between any two usages (either read or write) of the same variable, and \captionn{iii} \emph{returns-to}, for edges introduced between a return statement and the function declaration. All edges are initially added in both directions, but can be later removed during the training.
Depending on the task, more edge types can be easily~added.

\paragraph{Representation refinement}
Our goal is to learn an abstraction function $\alpha\colon \langle V, E, \xi_V, \xi_E \rangle\!\rightarrow\!\langle V, E'\!\subseteq\!E, \xi_V, \xi_E \rangle$ that removes a subset of the edges from the graph.
To quantify the size of the abstraction, we use $\abs{\alpha(x)} \coloneqq \abs{E'}$ to denote the number of edges after applying $\alpha$ on $x$.

\paragraph{Defining valid graph refinements}
Because the goal of representation refinement is to reduce the number of nodes on which a prediction depends, we need to ensure that $\alpha$ itself does not depend on all the graph nodes.
This is necessary as otherwise we only shift the dependency on the \emph{entire} program from the model $f$ to the representation refinement~$\alpha$.
To achieve this, the decision to include or remove a~given edge is done \emph{locally}, based only on the edge attributes and attributes of the nodes it connects.

Concretely, for a given edge $\langle s, t \rangle \in E$, we define an edge feature $\phi(\langle s, t \rangle) \coloneqq \langle \xi_E(\langle s, t \rangle), \xi_V(s), \xi_V(t) \rangle$ to be a tuple of the edge attributes and attributes of the nodes it connects.
As a form of regularization, we condition only on the \emph{type} attribute of each node.
We denote the set of all possible~edge features $\Phi$ to be the range of the function $\phi$ evaluated over all edges in $\dataset$.
Further, we define the refinement $\alpha$ as a subset of edge features $\alpha \subseteq \Phi$.
Finally, the semantics of executing $\alpha$ over edges $E$ is that only edges whose features are in $\alpha$ are kept, i.e., $\{ e \mid e \in E \wedge \phi(e) \in \alpha \}$.

\paragraph{Problem statement} Minimize the expected size of the refinement $\alpha \subseteq \Phi$ subject to the constraint that the expected loss of the model $f$ stays approximately the same:
\begin{equation}\label{refinement_statement}
\argmin_{\alpha \subseteq \Phi}  \sum_{(x, y) \in \dataset} \abs{\alpha(x)}
\end{equation}
\text{subject to}
\begin{equation*}
\textstyle \sum_{(x,y) \in \dataset} \ell(f(x), y) \approx \sum_{(x,y) \in \dataset} \ell(f(\alpha(x)), y)
\end{equation*}

Our problem statement is quite general and can be instantiated by: \captionn{i} using $\ell_{\scode{AbstainCrossEntropy}}$ as the loss (Appendix~B), and \captionn{ii} using \textit{adversarial risk} (Appendix C).

Allowing the model to abstain from making predictions is especially important in order to obtain small~$\alpha$ (i.e., sparse graphs). This is because the restriction that the model accuracy is roughly the same is otherwise too strict and would require that most edges are kept. Further, note that the problem formulation is defined over \emph{all} samples in $\dataset$, not only those for which the model $f$ predicts the correct label. This is necessary since the model needs to make a prediction for all samples, even if that prediction is to abstain.

\paragraph{Optimization via integer linear programming (ILP)}
To solve \Eqref{refinement_statement}, the key idea is that for each sample $(x, y)\!\in\!\dataset$ we first capture the relevance of each node to the prediction made by the model~$f$ by computing:
\begin{equation*}
\va(f, x, y) = \big[ \norm{\rmG_{i, :}}_1, \dots,  \norm{\rmG_{\abs{p}, :}}_1 \big],
%
\end{equation*}
where $\rmG = \nabla_{x}~\ell(f(x), y) \in \sR^{\abs{p} \times emb}$ denotes the gradient with respect to the input $x=\langle p, l \rangle$ and a given prediction $y$. As positions in $p$ correspond to discrete words, the gradient is computed with respect to their embedding $emb \in \sR$.
The score for each position in $p$ is computed by applying the $\normlone$-norm over the embedding gradients, producing a vector of unnormalized scores $\va \in \sR^{\abs{p}}$. To obtain a probability distribution $\hat{\va}(f, x, y)$ over all positions in $p$, we normalize the entries in $\va$ accordingly.

Then, we phrase the solution of \Eqref{refinement_statement} as an optimization problem of including the minimum number of edges necessary for a~path to exist between every relevant node (according to $\hat{\va}$) and the  node where the prediction is~made.
Preserving all paths between the prediction and relevant nodes encodes~the constraint that the expected loss stays approximately the~same, since it allows propagating information throughout the graph neural network.
This optimization can be naturally encoded as minimum-cost maximum-flow problem and solved efficiently with off-the-shelf ILP solvers.
We provide formal definition of the ILP encoding as well as concrete examples in Appendix~D.

Even though our ILP formulation is very fast (in all our experiments the ILP solver takes less than a second), it does result in a more complex approach compared to an end-to-end trainable solution.
We note however that an end-to-end trainable solution is also possible.
For example, one could make $\alpha$ continuous by defining a learnable weight for each edge feature $\phi$, encode the sparsity on $\alpha$ as part of the loss, and extend the graph neural network such that each message propagated along an edge $e$ is scaled using the corresponding value of the edge feature $\phi(e)$.
We have explored this option in the work of \cite{korbinian}.

\section{Training Algorithm}\label{training}
We now describe our algorithm that combines learning to abstain, adversarial training and representation refinement.

\begin{algorithm}[t]
 \caption{Training procedure used to learn a single adversarially robust model $\langle f, g_h, \alpha \rangle$.}
 \label{single_robust}
 \begin{algorithmic}[1]
 \FUNCTION {\scode{RobustTrain}$(\dataset, t_{\scode{acc}}):$}
 \STATE $\alpha_{\scode{last}} \gets \Phi$
 \STATE $f, g_h \gets$ \scode{Train} $(\dataset, t_{acc} - \epsilon)$
 \WHILE {$true$}
 \STATE $\alpha \gets$ \scode{RefineRepresentation} $(\dataset, f, g_h)$
 \STATE {\bfseries if} $\abs{\alpha} \geq \abs{\alpha_{\scode{last}}}$ {\bfseries then break}
 \STATE $\alpha_{\scode{last}} \gets \alpha$
 \STATE $\dataset \gets \{ (\alpha(x), y) \mid (x, y) \in \dataset \}$
 \STATE $f,\!g_h\!\gets\!$ \scode{AdversarialTrain} $(\dataset,\!f,\!g_h, t_{\scode{acc}} - \epsilon)$
 \ENDWHILE
 \STATE set threshold $h$ in $g_h$ such that the accuracy is $t_{\scode{acc}}$ 
 \STATE {\bfseries return} $\langle f, g_h, \alpha \rangle$
 \ENDFUNCTION
\end{algorithmic}
\end{algorithm}
\paragraph{Training a single adversarially robust model}
The training procedure used to learn a single adversarially robust model is shown in \Algref{single_robust}.
The input is a~training dataset $\dataset$ and the desired accuracy $t_{\scode{acc}}$ that the learned model should have.
Here, setting $t_{\scode{acc}}\!=\!1.0$ corresponds to a model that makes no mis-prediction (i.e., $100\%$ accuracy) while $t_{\scode{acc}}\!=\!0$ corresponds to training a~model that never abstains.

We start by training a~model $f$ and a~selection function $g_h$ as described in Appendix B (line 3).
At this point we do not use adversarial training and train with a weaker threshold $t_{\scode{acc}} - \epsilon$, as our goal is only to obtain a~fast approximation of the samples that can be predicted with high certainty.
We use $f$ and~$g_h$ to obtain an initial representation refinement~$\alpha$ (line~5) which is applied to the dataset $\dataset$ to remove edges that are not relevant according to $f$ and $g_h$ (line 8).
After that, we perform adversarial training (line 9) as described in Appendix C.
However, instead of training from scratch, we reuse model $f$ and $g_h$ learned so far, which speeds-up training.
Next, we refine the representation again (line 5) and if the new representation is smaller (line~6), we repeat the whole process.
Note that the adversarial training also uses threshold $t_{\scode{acc}} - \epsilon$ to account for the fact that the suitable representation is not known in advance.
After the training loop finishes, we set the threshold $h$ used by~$g_h$ to match the desired accuracy $t_{\scode{acc}}$ (more details on this step are provided in Appendix B).
The final result is a model consisting of the function $f$ trained to make adversarially robust predictions, the selection function~$g_h$ and the abstraction $\alpha$.

\vspace{-0.3em}
\paragraph{Incorporating robust predictions}
Once a single model is learned, it makes robust predictions on a~subset of the dataset $\dataset_{\scode{predict}}\!=\!\{ (x, y) \mid (x, y)\!\in\!\dataset \wedge g_h(\alpha(x))\!\geq\!h \}$~and abstains from making a prediction on the remainder of the samples $\dataset_{\scode{abstain}} = \dataset \setminus \dataset_{\scode{predict}}$.
Next, for all samples in $\dataset_{\scode{predict}}$, we use the learned model to annotate the position~$l$ in the program $p$ (recall that each $x =\langle p, l \rangle$ consists of a~program~$p$ and a~position $l$) with the ground-truth label~$y$ (denoted as $\scode{Apply}$ in \Algref{multi_robust}).
Annotating a~program position corresponds to either defining a new attribute (as illustrated in \Figref{overview}\textit{e}) or replacing an existing attribute (e.g., the \textit{value} attribute) of a given node. 
Note that annotating programs is useful only in cases where the same program $p$ is shared by multiple samples $(x, y) \in \dataset$ (i.e., multiple predictions are computed for different positions in the same program).

\begin{algorithm}[t]
 \caption{Training multiple adversarially robust models, each of which learns to make predictions for a different subset of the dataset~$\dataset$.}
 \label{multi_robust}
 \begin{algorithmic}[1]
 \FUNCTION {\scode{AccurateAndRobustTrain}$(\dataset,t_{\scode{acc}}\!=\!1.0)$}
 \STATE $M \gets []$
 \WHILE {$true$}
 \STATE $\langle f, g_h, \alpha \rangle \gets$ \scode{RobustTrain}$(\dataset, t_{\scode{acc}})$
 \STATE $\dataset_{\scode{abstain}} \gets$ \scode{Apply}$(\dataset, f, g_h, \alpha)$
 \STATE {\bfseries if} $\abs{\dataset_{\scode{abstain}}} = \abs{\dataset}$ {\bfseries then break}
 \STATE $\dataset \gets \dataset_{\scode{abstain}}$
 \STATE $M \gets M \cdot \langle f, g_h, \alpha \rangle $
 \ENDWHILE
 \STATE {\bfseries return} $M$
 \ENDFUNCTION
\end{algorithmic}
\end{algorithm}

\vspace{-0.3em}
\paragraph{Main training algorithm}
Our main training algorithm is shown in \Algref{multi_robust}. It takes as input the training dataset~$\dataset$ and learns multiple models $M$, each of which makes robust predictions on a different subset of $\dataset$ (as motivated in \Secref{overview}).
The number of models and the subsets for which they make predictions is not fixed a priori and is learned as part of~our training.
Model training (line~4) and model application (line~5) are performed as long as a~non-empty robust model exists (i.e., it makes at least one prediction).
If the goal is to make predictions for all the samples in~$\dataset$, the \Algref{multi_robust} is run iteratively, with decreasing values of $t_{\scode{acc}}$ until the full dataset is covered.

\vspace{-0.3em}
\paragraph{Verifying model correctness}
A natural extension of our approach is to formally verify that the learned models are correct.
Even though formally verifying the correctness of all samples is typically infeasible, it is possible to verify a~subset of them. This can be achieved since using representation refinement significantly simplifies the problem of proving correctness of \textit{all} positions (nodes) in the program to a much smaller set of relevant positions.
In fact, for some cases the refined representation is so small that it is possible to simply enumerate all valid modifications (e.g., a finite set of valid variable renamings) and check that the model is correct for all of them.
Additionally, it would be possible to adapt the recently proposed techniques~\cite{huang-2019, jia2019certified}, based on Interval Bound Propagation, that verify robustness to any valid word renaming and word substitution modifications.
However, applying these techniques to realistic networks in a scalable and precise ways is an open problem beyond the scope of our work.

\section{Evaluation}\label{eval}
We instantiated our approach to a well studied task -- predicting types for dynamically typed languages \scode{JavaScript} and \scode{TypeScript}~\cite{DeepTyper:18, Schrouff:2019, NL2Type:19, JSNice}.
In this task, the need for model robustness is natural since the model is queried each time a~program is modified by the user.
Our key results are:

\begin{itemize}
\item \textit{Our approach learns accurate and adversarially robust models} for the task of type inference, achieving $87.7\%$ accuracy while improving robustness from $52.1\%$ to $67.0\%$. 

\item We train highly accurate and robust models \textit{for a subset of the dataset}, with $99.9\%$ accuracy and $99.9\%$ robustness for $29\%$ of the samples. 


\end{itemize}

Our implementation uses PyTorch~\cite{pytorch} and DGL library~\cite{wang2019dgl}. We used a single Nvidia TITAN RTX for all the experiments. For our dataset, we collect the same top starred projects on Github and perform similar preprocessing steps as \citeauthor{DeepTyper:18} We provide detailed description in the supplementary material. The code and datasets are available at:

\begin{center}
\url{https://github.com/eth-sri/robust-code}
\end{center}

\paragraph{Evaluation metrics}
We use two main evaluation metrics:

\emph{Accuracy} is computed over the unmodified dataset $\dataset$ and corresponds to the accuracy used in prior works.
Concretely, the accuracy is defined as the ratio of samples $(x,y)$ for which the most likely label according to the model $f$, denoted $f(x)_{\scode{best}}$, is the same as the ground truth label $y$:
\begin{equation*}
\frac{1}{\abs{\dataset}} \sum_{(x,y) \in \dataset}
\begin{cases}
    1  & \quad \text{if } f(x)_{\scode{best}} = y\\
    0  & \quad \text{otherwise}
\end{cases}
\end{equation*}
\textit{Robustness} is the ratio of samples $(x, y) \in \dataset$ for which the model $f$ evaluated on all valid modifications $\mods \subseteq \Delta(x)$ either abstains or makes a correct prediction:
\begin{equation*}
\frac{1}{\abs{\dataset}} \sum_{(x,y) \in \dataset}
\begin{cases}
    0  & \text{if\;\;} \exists_{\mods \subseteq \Delta(x)} f(x + \mods)_{\scode{best}} \notin \{ y, \scode{abstain} \}\\
    1  & \text{otherwise}
\end{cases}
\end{equation*}

\paragraph{Models}
We evaluate five neural model architectures:


\lstm{} is a bidirectional LSTM with attention which takes as input a sequence of AST nodes, including both types and values, obtained using pre-order traversal.

\deeptyper{} is a model proposed by \citeauthor{DeepTyper:18} and consists of a bidirectional LSTM layer, followed by a single layer graph neural network that connects all variables with the same name (referred as consistency layer), followed by another bidirectional LSTM layer. Our only modification is that the input to our model is a sequence of AST types and values, instead of using syntactic program tokens.

\gcn{}, \ggnn{} and \gnt{} are three graph neural networks that use as input the graph program representation described in \Secref{refinement}. Here, \gcn{} is a Graph Convolutional Network~\cite{kipf2017semi}, \ggnn{} is Gated Graph Neural Network~\cite{li2016gated} and \gnt{} is a graph implementation of a~recently proposed transformer neural network architecture~\cite{transformer, dehghani2018universal}.
%

All models were trained with an embedding and hidden size of 128, batch size of 32, dropout~0.1~\cite{dropout}, initial learning rate of 0.001, using Adam optimizer~\cite{adam} and between 10 to 20 epochs.

\paragraph{Reducing dependencies via dynamic halting}
We further strengthen our \gnt{} model by implementing the
Adaptive Computation Time (ACT)~\cite{ACT} which dynamically learns how many computational steps each node requires in order to make a prediction.
This is in contrast to using a fixed amount of steps as done in~\cite{allamanis2018learning, brockschmidt2018generative}.
In our experiments, ACT significantly reduces the number of steps each node performs (half of the nodes perform $\leq\!3$ steps).

\begin{table*}[t]
\caption{Illustration of semantic and label preserving program modifications used in our work.}
\tablabel{program_modifications}
\tra{1.2}
\centering
\parbox{.38\linewidth}{
\footnotesize
\begin{tabular}{r c} 
\toprule

$\;$Substitutions and Renaming & Examples \\ 
\midrule
\multicolumn{2}{l}{\footnotesize \textbf{Semantic Preserving}}\\
variable renaming& \scode{x} $\rightarrow$ \scode{y} \\
object field renaming & \scode{obj.x} $\rightarrow$ \scode{obj.y}\\
property assignment renaming & \scode{\{x: obj\}} $\rightarrow$ \scode{\{y: obj\}} \\[0.3em]
\multicolumn{2}{l}{\footnotesize \textbf{Label Preserving}}\\
number substitution & \scode{2} $\rightarrow$ \scode{7} \\
string substitution & \scode{"get"} $\rightarrow$ \scode{"load"} \\
boolean substitution & \scode{true} $\rightarrow$ \scode{false} \\[1.0em]
\hline
\end{tabular}
}
\hfill
\parbox{.52\linewidth}{
\footnotesize
\begin{tabular}{r c} 
\toprule

$\;$Structural Modifications & Examples \\ 
\midrule
\multicolumn{2}{l}{\footnotesize \textbf{Label Preserving}}\\
new function parameters & \scode{def~inc(x)} $\rightarrow$ \scode{def~inc(x, y)}\\
new method arguments &  \scode{inc(x)} $\rightarrow$ \scode{inc(x, expr)}\\
\multicolumn{2}{l}{\footnotesize \textbf{Semantic Preserving}}\\
ternary expressions & \scode{expr}$_1$ $\rightarrow$ \scode{(expr)}$_2$ \scode{: expr}$_1$\scode{~?~expr}$_1$ \\
array access & \scode{expr} $\rightarrow$ \scode{[expr, expr][const]}\\
\multicolumn{2}{l}{\footnotesize \textbf{Dead Code}}\\
side-effect free expressions & $\emptyset \rightarrow$ \scode{expr}\\
adding object expressions & $\emptyset \rightarrow$ \scode{\{x: y, z: expr\}}\\
\hline
\end{tabular}
}

\end{table*}
\begin{table*}[t]
\caption{Comparison of accuracy and robustness across various models and training techniques considered in our work for the task of type inference. Adversarial training and the ability to abstain is applicable to all the models. The representation refinement is designed specifically to models defined over graphs, including \gcn{}, \ggnn{} and \gnt{}.}
\tablabel{baseline_accuracy}
\tra{1.2}
\parbox{.6\linewidth}{
\centering
\begin{tabular}{l c c c c c c} 
\toprule
& \multicolumn{2}{c}{\footnotesize \textbf{Standard Training}} & & \multicolumn{2}{c}{\footnotesize \textbf{Adversarial Training}}
\\
& \multicolumn{2}{c}{\footnotesize $\ell(f(x), y)$} && \multicolumn{2}{c}{\footnotesize $\max_{\mods \subseteq \Delta(x)} \ell(f(x + \mods), y)$}\\[0.3em]
$\;$Model & Accuracy & Robustness & & Accuracy & Robustness \\ 
\midrule
$\;\lstm$  		& $88.2\,\pm0.2$& $44.9\,\pm1.3$ &&  $87.5\,\pm0.4$	& $51.9\,\pm1.3$	\\ 	
$\;\deeptyper$ 	& $88.4\,\pm0.2$ & $52.4\,\pm1.2$ && $87.1\,\pm0.3$ & $55.1\,\pm2.6$ \\ 
$\,\gcn$ 		& $82.6\,\pm0.6$ & $49.1\,\pm1.1$ && $81.9\,\pm0.5$ & $49.3\,\pm3.1$	\\
$\,\gnt$ 		& $89.3\,\pm0.9$ & $47.4\,\pm1.0$ && $88.3\,\pm0.4$ & $50.0\,\pm0.5$	\\
$\,\ggnn$ 		& $86.7\,\pm0.4$ & $52.1\,\pm0.4$ && $86.1\,\pm0.2$ & $57.9\,\pm1.5$	\\

\hline
\end{tabular}

}
\hfill
\parbox{.38\linewidth}{
\centering
\begin{tabular}{l c c c c c c} 
\toprule
\multicolumn{3}{c}{\footnotesize \textbf{Abstain + Adversarial + Refinement}} 
\\
\multicolumn{3}{c}{\footnotesize $\max_{\mods \subseteq \Delta(x)} \ell_{\scode{AbstainCE}}((f, g_h)(\alpha(x + \mods)), y)$} \\[0.3em]
$\quad$Model & Accuracy & Robustness \\[-0.25em]
\midrule
\multicolumn{3}{l}{\footnotesize $t_\scode{acc} = 1.00$ (Abstain $\approx 70\%$)}\\[-0.2em]
\small $\quad\gnt$ 		& \small $99.93\%$ & \small $99.98\%$	\\[-0.25em]
\small $\quad\ggnn$ 	& \small $99.80\%$ & \small $99.01\%$	\\[-0.25em]
\footnotesize $t_\scode{acc} = 0.00$ \\[-0.2em]
\small $\quad\gnt$ 	& \small $86.6\%$ & \small $62.3\%$	\\[-0.25em]
\small $\quad\ggnn$ & \small $87.7\%$ & \small $67.0\%$	\\
\hline
\end{tabular}
}
\end{table*}

\paragraph{Program modifications}
We instantiate the adversarial training with both semantic preserving and label-preserving modifications shown in \Tabref{program_modifications}.
Here, $\scode{expr}$ is either an existing expression or a new expression consisting of a~random binary expression over constants up to depth 3, $\scode{const}$ is a randomly selected constant that results in a valid expression and $\scode{x, y, z}$ are variables in the program scope.
Our modifications extend those used by \citet{bielik2017learning} but the list is not exhaustive and can be extended further.

To measure the model robustness, we run the adversarial attack for 1000 iterations for renaming modifications and additional 300 iterations for structural modifications. These thresholds are rather high and were selected with the goal of closely estimating the true number of adversarial samples.
Further, note that since $\mods \subseteq \Delta(x)$ is a set, each iteration explores a set of concrete program modifications.

\subsection{Accurate and Adversarially Robust Models}

We summarize the main results in \Tabref{baseline_accuracy}.
The first column (left) shows the median test \textit{accuracy} and standard deviation of various models (across three trials trained with different random seeds).
The \gcn{} achieves the worst accuracy of $82.6\%$ and the accuracy of the remaining models is similar with \gnt{} model performing the best with $89.3\%$.


\paragraph{Existing models are not robust} 
While highly accurate, all models are also non-robust for up to half of the samples in the dataset.
In other words, for every second sample $x$ in our dataset, there exists a modification $\mods \subseteq \Delta(x)$ for which $f(x)$ predicts the type correctly while $f(x + \mods)$ mis-predicts it.
However, since these models were not trained with the goal of adversarial robustness, it is expected for them to be (atleast partially) non-robust.


\paragraph{Adversarial training alone is insufficient}
To improve the robustness, we next train the models using adversarial training as described in Appendix C.
Unfortunately, while the adversarial training increase the robustness, it does so only slightly.
The best improvement was achieved for \lstm{} and \ggnn{} models ($7\%$ and $5.8\%$, respectively). For \deeptyper{} and \gnt{} the robustness increased by $\approx 2.5\%$ while for \gcn{} it is only $0.2\%$.
This illustrates that while useful, if used alone, \textit{adversarial training} is not enough.


\paragraph{Our work: training accurate models with abstain}
The models trained using our approach are shown in \Tabref{baseline_accuracy} (right).
First, we trained our models to be both \textit{accurate} and \textit{robust} on a subset of the dataset. 
This can be achieved by setting the desired accuracy thresholds, in our case $t_{acc}\!=\!1.00$, which corresponds to training the model to make only correct predictions.
For $t_{acc}\!=\!1.00$, our approach learns an almost perfect model that is both accurate and robust for $\approx\!30.0\%$ of samples.
Here, \gnt{} learned 7 models and achieved $99.98\%$ robustness while \ggnn{} learned 8 models with robustness of $99.01\%$.
Learning multiple models is crucial for achieving higher coverage as a~single model would not abstain for only $17-20\%$ of the samples, compared to $30\%$ using multiple models.

The model did not achieve $100\%$ accuracy and robustness for $t_{acc}\!=\!1.00$ due to several samples included in the test set.
These samples were mis-predicted because they contained code structure not seen during training and not covered by modifications $\mods \subseteq \Delta(x)$.
This illustrates that it is important that the samples in $\dataset$ are diverse and contain all the language features and corner cases of the programs, or that the modifications $\Delta(x)$ are expressive enough such that these can be discovered automatically during training.

\paragraph{Our work: improving robustness}
Next, we train models that take advantage of the highly accurate and robust models trained using $t_{acc}\!=\!1.00$, but make predictions for all the samples (i.e., do not abstain).
This can be achieved by continuing the training while reducing $t_{acc}$ to zero and conditioning on all the models trained with higher $t_{acc}$.
In our experiments, we train a single additional model by directly setting $t_{acc}=0$ after training with $t_{acc}\!=\!1.00$.
The results are shown in \Tabref{baseline_accuracy} (right) and lead to additional robustness increase of $9.2\%$ and $12.3\%$ compared to using adversarial training only for \ggnn{} and \gnt{}, respectively.
For \gnt{}, the accuracy slightly decreases by $1.7\%$ which is expected as increasing model robustness typically comes at the~cost of reduced accuracy~\cite{tsipras2018robustness}.
Interestingly, for \ggnn{} our robust training increases the accuracy over both the adversarial training as well as standard training by $1.9\%$ and $1.0\%$, respectively.

\begin{table}[t]
\caption{Robustness breakdown for the \gnt{} and \ggnn{} models trained using our approach from \Tabref{baseline_accuracy} (right).}
\tablabel{breakdown}
\tra{1.2}
\begin{tabular}{l c c c c} 
\toprule
 & & \multicolumn{3}{c}{Robustness}\\
\cmidrule(r){3-5}
$\;$Dataset & Size & $\forall$ Correct & $\exists$ Incorrect &  Abstain \\ 
\midrule
\footnotesize \gnt{} & \multicolumn{4}{l}{\footnotesize $t_\scode{acc}=1.00\;\;$ } \\
$\;\dataset_{\scode{correct}}$ & $29.3\%$ & $90.0\%$ & $0.00\%$ & $10.00\%$ \\
$\;\dataset_{\scode{abstain}}$ & $70.6\%$ & $-$ & $0.01\%$ & $99.99\%$ \\
\footnotesize \ggnn{} &\multicolumn{4}{l}{\footnotesize $t_\scode{acc}=1.00$} \\
$\;\dataset_{\scode{correct}}$ & $30.6\%$ & $75.5\%$ & $0.06\%$ & $23.94\%$ \\
$\;\dataset_{\scode{abstain}}$ & $69.3\%$ & $-$ & $1.46\%$ & $98.54\%$ \\
\bottomrule
\multicolumn{5}{l}{\footnotesize $\forall$ Correct $\;\,\,\coloneqq \forall_{\mods \subseteq \Delta(x)} (f, g_h)(\alpha(x + \mods))_{\scode{best}} = y$ }\\[-0.2em]
\multicolumn{5}{l}{\footnotesize $\exists$ Incorrect $\coloneqq \exists_{\mods \subseteq \Delta(x)} (f, g_h)(\alpha(x + \mods))_{\scode{best}} \notin \{y, \scode{abstain} \}$ }
\end{tabular}
\end{table}

\paragraph{Adversarial robustness breakdown}
\Tabref{breakdown} provides a~detailed breakdown of the \emph{robustness} metric for the \gnt{} and \ggnn{} models trained with $t_{acc}\!=\!1.00$ from \Tabref{baseline_accuracy} (right).
Here, $\dataset_{\scode{abstain}}$ contains samples for which the model abstains from making a prediction and $\dataset_{\scode{correct}}$ contains samples for which the model evaluated on a non-adversarial input (i.e., $x$ without any modification) makes a~correct prediction.
We use $\forall$ correct to denote that a sample $(x,y)$ is correct for \textit{all} possible modifications $\mods\!\subseteq\!\Delta(x)$, the $\exists$ incorrect has the same definition as robustness (i.e., there exists a modification that leads to an incorrect prediction), and abstain denotes the remaining samples.

The \gnt{} is precise and keeps predicting the correct label in $90\%$ of cases and abstain in the rest.
This is even though the requirements for $\forall$ correct are very strict and require that all samples are correct.
When considering $\dataset_{\scode{abstain}}$, the \gnt{} model is also precise and produces incorrect prediction for only a single sample ($0.01\%$).
For \ggnn{} the results are similar but the model is both less precise (keeps the correct prediction in $75.5\%$ of cases) and less robust ($1.46\%$ of samples in $\dataset_{\scode{abstain}}$ can be modified to cause a~mis-prediction).
This shows that the majority of robustness errors from \Tabref{baseline_accuracy} are~due to mis-predicted samples for which the model originally abstained.


\section{Related Work}
Our work is related to a number of different areas from adversarial machine learning and learning over code.

\paragraph{Model certainty}
Several approaches have been recently proposed to extend neural models with certainty measure~\cite{gal16, DeepGamblers:19, galthesis, Geifman:2017, geifman19a}.
In our work, we use the method proposed by \citet{DeepGamblers:19} but in a novel way -- applied to the adversarial setting with the goal of training robust models.

\paragraph{Learning static analyzers from data}
A closely related work addresses the task of learning static analyzers~\cite{bielik2017learning}: it defines a domain specific language (DSL) to represent static analyzers, uses decision tree learning to obtain an interpretable model, and defines a procedure that finds counter-examples the model mis-classifies (used to re-train the model). At a high-level, some of the steps are similar but the actual technical solution is very different as we address a general class of neural models and do not assume any prior knowledge (i.e., a DSL).

\paragraph{Adversarial training}
Even though the problem of adversarial robustness of code has been overlooked, the adversarial training has already been applied to related domains -- natural language processing~\cite{shen2018deep, PapernotMSH16, Gao:2018, LiangLSBLS17, Belinkov:2017, hotflip} and graphs~\cite{dai18b, Zugner:2018, zugner2018adversarial}.

In the domain of \textit{graphs}, existing works focus on attacking the graph structure \cite{dai18b, Zugner:2018, zugner2018adversarial} by considering that the nodes are fixed and edges can be added or removed.
While this setting is natural for modelling many types of graphs, such approaches do not apply for the domain of code where graph edges can not be added and removed arbitrarily.

In \textit{natural language processing}, existing approaches generally: \captionn{i} measure the contribution of individual words or characters to the prediction (e.g., using gradients~\cite{LiangLSBLS17}, forward derivatives~\cite{PapernotMSH16} or head/tail scores~\cite{Gao:2018}), and \captionn{ii} replace or remove those whose contribution is high (e.g., using dictionaries~\cite{jia2019certified}, character level typos~\cite{Gao:2018, Belinkov:2017, hotflip}, or handcrafted strategies~\cite{LiangLSBLS17}).
The adversarial training used in our work operates similarly except our modifications are designed over programs.


\paragraph{Program representations}
A core challenge of using machine learning for code is designing a suitable program representation used as model input.
Due to its simplicity, the most commonly used program representation is a sequence of words, obtained either by tokenizing the program~\cite{DeepTyper:18} or by linearizing the abstract syntax tree~\cite{Li:2018}.
This however ignores the fact that programs do have a rich structure -- an issue addressed by representing programs as graphs~\cite{allamanis2018learning, brockschmidt2018generative} or as a~combination of abstract syntax tree paths~\cite{Alon:2019}.
In our work, we follow the approach proposed in recent works and represent programs as graphs. 
More importantly, we develop a novel technique that learns to refine the representation based on model predictions instead of fixing it a priori.
As shown in our evaluation,~this is crucial for learning robust models. 


\paragraph{Adversarial attacks for code}
Concurrent to our work, \citet{yefet2019adversarial} explored the task of generating adversarial attacks for code via gradient based optimization.
In contrast, we introduce an approach to reduce the search space an adversarial attack needs to consider by learning to refine the representation. Such reduced search space is useful for both for renaming and structural modifications, whereas gradient based optimization has been explored only for renaming.
However, both of these approaches are orthogonal and can be combined into one that learns both to reduce the search space as well as to efficiently find adversarial examples in this reduced search space.

\paragraph{Type inference}
We evaluated our work on the task of type inference for which a number of recent works improve accuracy by proposing a new neural architectures. 
In contrast, the goal of our work is to study and improve robustness of these models. 
To achieve this, we compare to two prior works in our evaluation~\cite{Schrouff:2019, DeepTyper:18}.
In addition to predicting types from source code, \citet{NL2Type:19} showed that it is possible to predict parameter types using natural language information obtained from method docstrings. 
Here, existing attacks on text (LSTM) can be used to assess its robustness but evaluating text models is outside the scope of our work.
Finally, two concurrent works to ours have proposed new models to improve accuracy: \scode{Typilus}~\cite{Typilus} and \scode{LambdaNet}~\cite{Wei2020LambdaNet}. Both of these works represent programs as graphs and use graph neural networks as the underlying model architecture, which makes our approach applicable.
However, we note that for \scode{LambdaNet} we expect the model to be quite robust as the authors manually designed a sparse graph representation (by designing a static analysis to extract the type dependence graph) over which to learn.

\section{Conclusion}
We presented a new technique to train \emph{accurate} and \emph{robust} neural models of code. Our work addresses two key challenges inherent to the domain of code: the difficulty of computing the correct label for all samples (i.e., the input is incomplete code snippet, program semantics are unknown) as well as the fact that programs are significantly larger and more structured compared to images or natural language.

To address the first challenge, we allow the model to abstain from making a prediction, rather than forcing the model to make predictions for all samples (as done in prior works).  
To address the second challenge, we learn which parts of the program are relevant for the prediction, and abstract~the rest (instead of using the \emph{entire} program as input). 

Further, we introduce a new procedure that trains multiple models, instead of one. This has several advantages, as each model is simpler and thus easier to train robustly, the learned representation is specialized to the kind of predictions it makes, and the model directly conditions on predictions of prior models (instead of having to re-learn them).
However, a disadvantage of our approach is that the models are learned sequentially which slows down the training (i.e., training $10$ models will take $10 \times$ more time). To speed up the training, it would be interesting to allow learning multiple models in parallel at each sequential step and then combine them as explored by \citet{ShazeerMMDLHD17}.

We believe than our work is only one step in addressing the task of adversarially robust models of code and that many challenges remain open.
For example, it remains to be seen how effective our approach is at other tasks over code, beyond type inference.
Further, we optimize for the worst case adversarial robustness, which corresponds to learning a~robust model for all programs. An interesting future work~is to optimize with respect to those modification that are common among developers, especially if it is not possible to be robust for all of them.
While we checked robustness~for a wide range of program modifications, these are still far from exhaustive and more work is needed in defining new ones.
Finally, as the number of possible modification is large and growing, an interesting area is designing how they can be combined efficiently, as explored recently by \citet{ramakrishnan2020semantic} and \citet{zhang2020robustness}.


\section*{Acknowledgements}
We would like to thank the anonymous reviewers who gave useful comments and provided interesting suggestions on how our work can be improved and extended.
Further, we would like to acknowledge the work of \cite{DeepTyper:18} which is publicly available and provided useful infrastructure for generating datasets used in our work.
The research leading to these results was partially supported by an ERC Starting Grant 680358.


\def\UrlBreaks{\do\/\do.}
\balance
\bibliography{bib}
\bibliographystyle{icml2020}


\cleardoublepage

\section*{Supplementary Material}
\appendix

We provide the following four appendices:
\begin{itemize}
\item \Appref{eval} provides details of our dataset and additional experiments that evaluates the effect of dataset size.

\item \Appref{uncertainty} describes the method (introduced by Liu et. al. \citeyear{DeepGamblers:19}) used in our work for training neural models that abstain from making predictions if uncertain.

\item \Appref{adversarial} describes application of the adversarial training in the domain of code via a set of program mutations.

\item \Appref{ilp} provides a formal definition of the integer linear encoding used to solve the problem in Equation~2 efficiently. Additionally, we provide a concrete example illustrating the encoding.
\end{itemize}

\section{Evaluation}\label{eval}

\paragraph{Implementation}
All our models are implemented in PyTorch~\cite{pytorch}. The graph neural networks are implemented using the DGL library v0.4.3~\cite{wang2019dgl}. To solve the integer linear program we use Gurobi solver v8.11~\cite{gurobi}. 



\paragraph{Adaptive computation time (ACT)~\cite{ACT}}
Our \gnt{} model implements the Adaptive Computation Time (ACT)~\cite{ACT} technique which dynamically learns how many computational steps each node requires in order to make a prediction.
This is instead of using a~fixed amount of steps as done for example in~\cite{allamanis2018learning, brockschmidt2018generative}.
To achieve this, recall that for each node $v_i \in V$ in the graph, a graph neural network computes sequence of hidden states $\vs^{i}_t$ where $t \in \sN$ is the timestep\footnote{
Note we assume only that $\vs^{i}_t$ is computed for each timestep which is independent of the concrete graph neural architecture used to compute~$\vs^{i}_t$.}.
Following \cite{ACT}, the number of timesteps that the model performs is controlled by introducing an extra sigmoidal halting unit $h^i_t \in \sR^{(0,1)}$ with associated learnable weight matrix~$W_h$ and bias $b_h$:
\[
h^i_t = \sigma(W_h s^i_t + b_h)
\]
The output of the halting unit is then used to determine the halting probability $p^i_t$ as follows:
\[
p^i_t = \begin{cases}
1 - \sum_{k=0}^{t - 1} h^i_k & \text{if~} t = T \text{~(last timestep)}\\
1 - \sum_{k=0}^{t - 1} h^i_k & \text{if~} \sum_{k=0}^{t} h^i_k \geq 1 - \epsilon\\
h^i_t & \text{otherwise}\\
\end{cases}
\]
where $T \in \sN$ is the maximum allowed number of timesteps and
$\epsilon \in \sR^{(0,1)}$ is a small constant introduced to allow the network to stop after a single step (we use $\epsilon = 0.01$ in our experiments). 
Finally, the halting probability $p^i_t$ is used to define the final state $\vs^i_T$ of a node $v_i$ as a weighted average of its intermediate states:
\[
\vs^i_T = \sum_{t=0}^T p^i_t \cdot \vs^i_t
\]

\paragraph{Dataset}
To obtain the datasets used in our work, we extend the infrastructure from \scode{DeepTyper}~\cite{DeepTyper:18}, collect the same top starred projects on GitHub, and perform similar preprocessing steps -- remove \scode{TypeScript} header files, remove files with less than 100 or more than 3,000 tokens and split the projects into train, validation and test datasets such that each project is fully contained in one of the datasets.
Additionally, we remove exact file duplicates and files that are similar to each other ($\approx\!10\%$ of the files). We measure file similarity by collecting all 3-grams (excluding comments and whitespace) and removing files with Jaccard similarity greater than~$0.7$.

We compute the ground-truth types using the \scode{TypeScript} compiler version $3.4.5$ based on manual type annotations, library specifications and analyzing all project files.
While we reuse the same GitHub projects and part of \scode{DeepTyper}'s infrastructure\footnote{\url{https://github.com/DeepTyper/DeepTyper}} to obtain the dataset, the datasets are not directly comparable for a number of reasons.
First, we fixed a bug due to which \scode{DeepTyper} incorrectly included some type annotations as part of the input. 
Second, the projects we used are subset of those used in \scode{DeepTyper} since some are no longer available and were removed from GitHub.
Third, we additionally predict the types corresponding to all intermediate expressions and constants (e.g., the expression $x + y$ contains three predictions for $x$, $y$ and $x+y$).
This improves model performance as it is explicitly trained also on the intermediate steps required to infer the types.
Finally, we train all the models to predict four primitive types (\scode{string}, \scode{number}, \scode{boolean}, \scode{void}), four function types ($()\!\Rightarrow\!\scode{string}$, $()\!\Rightarrow\!\scode{number}$, $()\!\Rightarrow\!\scode{boolean}$, $()\!\Rightarrow\!\scode{void}$) and a~special \scode{unk} label denoting all the other~types.
While this is similar to types predicted by some other works such as \scode{JSNice}~\cite{JSNice}, it is only subset of types considered in \scode{DeepTyper}.


All results presented in Tables 1 and 2 are obtained by training our models using a dataset that contains 3000 programs split equally between training, validation and test datasets.
Because each program contain multiple type predictions, the number of training samples is significantly higher than the number of programs. Concretely, there are $139,915$, $223,912$ and $121,153$ samples in training, validation and test datasets.
We note that this is only a subset of the full dataset that can be obtained by processing all the files included in the projects used by \citeauthor{DeepTyper:18} We make the dataset available online at \url{https://github.com/eth-sri/robust-code}.

During adversarial training, we explore 20 different modifications $\mods \subseteq \Delta(x)$ applied to each sample~$(x, y) \in \dataset$ which effectively increases dataset size by up to two orders of magnitude since for each training epoch the modifications are different.
For the purposes of evaluation, we increase the number of explored modifications to 1300 for each sample -- 1000 for renaming modifications and further 300 for renaming together with structural modifications.

%
\section{Training Neural Models to Abstain}\label{uncertainty}

We now present a method for training neural models of code that provide an uncertainty measure and can abstain from making predictions. This is important as essentially~all practical tasks contain some examples for which it is not possible to make a correct prediction (e.g., due to the task~hardness or because it contains ambiguities). In the machine learning literature this problem is known as selective classification (supervised-learning with a reject option) and is an active area with several recently proposed approaches~\cite{gal16, DeepGamblers:19, galthesis, Geifman:2017, geifman19a}. 
In our work, we use one of these methods \cite{DeepGamblers:19} which is briefly summarized below. For a full description, we refer the reader to the original paper \cite{DeepGamblers:19}.

Let $\dataset\!=\!\{ (x_j, y_j) \}_{j=1}^N$ be a training dataset and $f\colon \sX\!\rightarrow\!\sY$ an existing model trained to make predictions on $\dataset$.
The existing model~$f$ is augmented with an option to abstain from making a~prediction by introducing a~selection function $g_h\colon \sX~\rightarrow~\sR^{(0, 1)}$ with an associated threshold $h \in \sR^{(0, 1)}$, which leads to the following definition:
\begin{equation}
(f, g_h)(x) \coloneqq
\begin{cases}
   \;\; f(x)      & \quad \text{if } g_h(x) \geq h\\
   \;\; \texttt{abstain}  & \quad \text{otherwise}
\end{cases}
\end{equation}
That is, the model makes a prediction only if the selection function $g_h$ is confident enough (i.e., $g_h(x) \geq h$) and abstains from making a prediction otherwise.
Although conceptually the model is now defined by two functions $f$ (the original model) and $g_h$ (the selection function), it is possible to adapt the original classification problem such that a single function $f'$ encodes both.
To achieve this, an additional \scode{abstain} label is introduced and a function $f'\colon \sX\!\rightarrow\!\sY \cup \{\scode{abstain}\}$ is trained in the same~way as~$f$ (i.e., same network architecture, hyper-parameters, etc.) with two exceptions: \captionn{i} $f'$ is allowed to predict the additional \scode{abstain} label, and \captionn{ii} the loss function used to train $f'$ is changed to account for the additional label.
After $f'$ is obtained, the selection function is defined as $g_h \coloneqq 1 - f'(x)_{\scode{abstain}}$, that is, to be the probability of selecting any label other than \scode{abstain} according to $f'$.
Then, $f$ is defined to be re-normalized probability distribution obtained by taking the distribution produced by $f'$ and assigning zero probability to \scode{abstain} label.
Essentially, as long as there is sufficient probability mass $h$ on labels outside \scode{abstain}, $f$ decides to select one of these labels.

\paragraph{Loss function for abstaining}
To gain an intuition behind the loss function used for training $f'$, recall that the standard way to train neural networks is to use cross entropy~loss:
\begin{equation}
\ell_{\scode{CrossEntropy}} (\vp, \vy) \coloneqq - \sum_{i=1}^{\abs{\sY}} y_i \log(p_i)
\end{equation}
Here, for a given sample $(x,y) \in \dataset$, $\vp = f(x)$ is a vector of probabilities for each of the $\abs{\sY}$ classes computed by the model and $\vy\!\in\!\R^{\abs{\sY}}$ is a vector of ground-truth probabilities.
Without loss of generality, assume only a~single label is correct, in which case $\vy$ is a~one-hot vector (i.e., $y_j\!=\!1$ if $j$-th label is correct and zero elsewhere).
Then, the cross entropy loss for an example where the \textit{j}-label is correct is $-\log(p_j)$. Further, the loss is zero if the computed probability is $p_j = 1$ (i.e., $-\log(1) = 0$) and positive otherwise.

Now, to incorporate the additional \scode{abstain} label, the abstain cross entropy loss is defined as follows:
\begin{equation}\label{rejection_ce}
\ell_{\scode{AbstainCrossEntropy}} (\vp, \vy) \coloneqq - \sum_{i=1}^{\abs{\sY}} y_i \log(p_i o_i + p_{\scode{abstain}})
\end{equation}
Here $\vp \in \sR^{\abs{\sY} + 1}$ is a distribution over the classes (including \scode{abstain}), $o_i \in \sR$ is a constant denoting the weight of the \textit{i}-th label and $p_{\scode{abstain}}$ is the probability assigned to \scode{abstain}. Intuitively, the model either: \captionn{i} learns to make ``safe'' predictions by assigning the probability mass to $p_\scode{abstain}$, in which case it incurs constant loss of $p_\scode{abstain}$, or \captionn{ii} tries to predict the correct label, in which case it potentially incurs smaller loss if $p_i o_i > p_{\scode{abstain}}$. If the scaling constant $o_i$ is high, the model is encouraged to make predictions even if it is uncertain and potentially makes lot of mistakes. As $o_i$ decreases, the model is penalized more and more for making mis-predictions and learns to make ``safer'' decisions by allocating more probability mass to the \scode{abstain} label.

\paragraph{Obtaining a model which never mis-predicts on $\dataset$}
For the $\ell_{\scode{AbstainCrossEntropy}}$ loss, it is possible to always obtain a model $f'$ that never mis-predicts on samples in $\dataset$. Such a model $f'$ corresponds to minimizing the loss incurred by \Eqref{rejection_ce} which corresponds to maximizing $p_i o_i + p_{\scode{abstain}}$ (assuming $i$ is the correct label). This can be simplified and bounded from above to $p_i\!+\!p_{\scode{abstain}}\!\leq\!1$, by setting $o_i\!=\!1$ and for any valid distribution it holds that $1\!=\!\sum_{p_i \in \vp} p_i$. Thus, $p_i o_i + p_{\scode{abstain}}$ has a global optimum trivially obtained if $p_\scode{abstain}\!=\!1$ for~all samples in $\dataset$. That is, the correctness (no mis-predictions) can be achieved by rejecting all samples in~$\dataset$. However, this leads to zero recall and is not practically useful.

\paragraph{Balancing correctness and recall}
To achieve both correctness and high recall, similar to Liu et. al., we train our models using a form of annealing. We start with a high $o_i=\abs{\sY}$, biasing the model away from abstaining, and then train for a number of epochs~$n$. We then gradually decrease $o_i$ to 1 for a fixed number of epochs $k$, slowly nudging it towards abstaining. Finally, we keep training with $o_i=1$ until convergence. We note that the threshold $h$ is not used during the training. Instead, it is set after the model is trained and is used to fine-tune the trade-off between recall and correctness (precision).
Further, note that $o_i=1$ is used only if the desired accuracy is $100\%$ and otherwise we use $o_i=1+\epsilon$. Here, $\epsilon$ is selected by decreasing the value $o_i$ as before but stopping just before the model abstains from making all predictions.

\paragraph{Summary}
We described an existing technique~\cite{DeepGamblers:19} for training a model that learns to abstain from making predictions, allowing us to trade-off correctness (precision) and recall. A key advantage of this technique is its generality -- it works with any existing neural model with two simple changes: \captionn{i} adding an \scode{abstain} label, and \captionn{ii} using the loss function in \Eqref{rejection_ce}. To remove clutter and keep discussion general, the rest of our work interchangeably uses $f(x)$ and $(f, g_h)(x)$.

\section{Adversarial Training for Code}\label{adversarial}
In \Secref{uncertainty}, we described how to learn models that are correct on subset of the training dataset $\dataset$ by allowing the model to abstain from making a prediction when uncertain.
We now discuss how to achieve robustness (that is, the model either abstains or makes a correct prediction) to a~much larger (potentially infinite) set of samples beyond those included in $\dataset$ via so-called \textit{adversarial training}~\cite{GoodfellowSS14}.

\paragraph{Adversarial training}
The goal of adversarial training~\cite{madry2018towards, wong18a, sinha2018certifiable, raghunathan2018certified} is to minimize the expected adversarial loss:

\begin{equation}\label{adversarial_expected_loss}
\E_{(x, y) \sim D}[ \max_{\mods \subseteq \Delta(x)} \ell(f(x + \mods), y)]
\end{equation}

In practice, as we have no access to the underlying distribution but only to the dataset $\dataset$, the expected adversarial loss is approximated by \textit{adversarial risk} (which training aims to minimize):
\begin{equation}\label{adv_risk}
\dfrac{1}{|\dataset|}\sum_{(x, y) \in \dataset}^{\abs{\dataset}} \max_{\mods \subseteq \Delta(x)} \ell(f(x + \mods), y)
\end{equation}

Intuitively, instead of training on the original samples in~$\dataset$, we train on the worst perturbation of each sample. Here, $\mods \subseteq \Delta(x)$ denotes an ordered sequence of modifications while $x + \mods$ denotes a new input obtained by applying each modification $\delta \in \mods$ to $x$. Recall that each input $x=\langle p, l \rangle$ is a tuple of a program $p$ and a position $l$ in that program for which we will make a prediction. Applying a modification $\delta\colon \sX\!\rightarrow\!\sX$ to an input $x$ corresponds to generating both a~new program as well as updating the position $l$ if needed (e.g., in case the modification inserted or reordered program statements). That is,
$\delta$ can modify \emph{all} positions in $p$, not only those for which a prediction is made. Further, note that the sequence of modifications $\mods \subseteq \Delta(x)$ is computed for each $x$ separately, rather than having the same set of modifications applied to all samples in $\dataset$. 

Using adversarial training in the domain of code requires a set of \emph{label preserving} modifications $\Delta(x)$ over programs which preserve the output label $y$ (defined for a given task at hand), and a~technique to solve the optimization~problem $\max_{\mods \subseteq \Delta(x)}$ efficiently. We elaborate on both of these next.

\subsection{Label Preserving Program Modifications}
We define three types of label preserving program modifications -- word substitutions, word renaming, and sequence substitutions.
Note that label preserving modifications are a~strict superset of semantic preserving modifications. 
This is because while label preserving modifications only require that the correct label does not change, the semantic preserving modifications require that both the label does not change as well as that the overall program semantics do not change. 
Preserving programs semantics is for many properties unnecessarily strict and therefore we focus on the more general label preserving modifications.

\begin{itemize}
\item \textit{Word substitutions} are allowed to substitute a~word at a single position in the program with another word (not necessarily contained in the program).
Examples of word substitutions include changing constants or values of binary/unary operators.

\item \textit{Word renaming} is a~modification which includes renaming variables, parameters, fields or methods.
In order to produce valid programs, this modification needs to ensure that the declaration and all usages are replaced jointly. Because of this, renaming a single variable in practice always corresponds to making multiple changes to the program (i.e., $\abs{\delta} > 1$ unless the variable is used only once).

\item \textit{Sequence substitution} is the most general type of modification which can perform any label preserving program change such as adding dead code or reordering independent program statements.
\end{itemize}

The main property differentiating the modification types is that word renaming and substitution do not change program structure.
This is used both to compute which substitution should be made as well as to~provide formal correctness guarantees (discussed in Section 4).
Further, is it used for efficient implementation that allows us to implement word substitutions and word renaming directly on the batched tensors, thus making them fast. In contrast, sequence substitutions require parsing batched tensors back to programs, applying modifications on the programs and the processing the resulting programs back to batched tensors. As a result, word substitutions and renaming take $0.1$ second to apply once over the full training dataset while structural modifications are $\approx 70\times$ slower and take $7$ seconds.

Additionally, it is also possible to define modifications that are not label preserving (i.e., change the ground-truth label), in which case the user has to additionally provide an oracle that computes the correct label $y$.
However, such oracles are typically expensive to design and run (i.e., one would need to run a static analysis over the program or execute the program) and therefore label preserving modification are a~preferred option whenever available.

\subsection{Finding Adversarial Examples}

Given a program $x$, associated ground-truth label $y$, and a~set of valid modifications $\Delta(x)$ that can be applied over~$x$, our goal is to select a subset of them $\mods \subseteq \Delta(x)$ such that the inner term in the adversarial risk formula $\max_{\mods \subseteq \Delta(x)} \ell(f(x + \mods), y)$ is maximized.
Solving for the optimal $\mods$ is highly non-trivial since: \captionn{i} $\mods$ is an ordered sequence rather than a single modification, \captionn{ii} the set of valid modifications $\Delta(x)$ is typically very large, and \captionn{iii} the modification can potentially perform arbitrary rewrites of the program (due to sequence substitutions). Thus, we focus on solving this maximization approximately, inline with how it is solved in other domains.
In what follows, we discuss three approximate approaches to achieve this and discuss their advantages and limitations.

\subsubsection{Greedy Search}

The first approach is a greedy search that randomly samples a sequence of modifications $\mods \subseteq \Delta(x)$. The sampling can be performed for a predefined number of steps with the goal of maximizing the adversarial risk, or until an adversarial example is found (i.e., $f(x + \mods) \neq f(x)$). 
Concretely, for a given input $x = \langle p, l \rangle$, let us define the space of valid modifications $\Delta(x) \subseteq \Delta(p, l_1) \times \Delta(p, l_2) \times \dots \times \Delta_n(p, l_n)$ as the Cartesian product of possible modification applied to each position in the program $l_{1:n}$.
We select $\mods$ using the following procedure: sample a threshold value $t \sim \mathcal{N}(0.1, 0.4)$ and apply the modification at each location with probability $t$. If $\abs{\Delta_i(p, l_i)} > 1$, then the modification to apply is sampled at random from the set $\Delta_i(p, l_i)$.
Sampling of the threshold value $t$ is done per each sample $x$ and ensures variety in the number of modifications applied.

\paragraph{Limitations and advantages}
The main advantage of this technique is that it is simple, easy to implement, and very fast.
Given its simplicity, this technique is independent of the actual modification and applies equally to words substitutions, word renamings as well as sequence substitutions.
However, a natural limitation of this technique is that is uses no information about which positions and which values are important to the prediction is used.

\subsubsection{Gradient-based Search}
Similar to prior works, gradient information can be used to guide the search for an adversarial examples.
This can done in two ways -- \captionn{i} finding a program position to change, and \captionn{ii} finding both a program positions as well as the new value to change.
To find a program position, we can use gradients to measure the importance of each position $\va$ for a given prediction in the same way as described in Section~3.
Once the \textit{attribution} score $\va$ is computed, the adversarial attack can be generated by sampling positions to be modified proportionally to $\va$, instead of the uniform sampling used in the greedy search.

Additionally, as shown in the concurrent work~\cite{yefet2019adversarial}, the gradients can also be used to select both the program position and the new value to be used (instead of sampling from all valid values uniformly at random).

\textbf{Limitations and advantages}
The main advantage of gradient-based approach is that the decision of which position to changes as well as what the new value should be is guided, rather than random.
Further, for renaming modifications, such approach has shown to be quite effective \cite{yefet2019adversarial} at finding the adversarial examples.
However, the main limitation of this approach is that it works only for replacing single value (i.e., word substitutions and word renaming) and not when the value is a~complex structure (i.e., sequence substitution).
Sequence substitutions are important class of modifications which are however hard to optimize for as in general, they can perform arbitrary changes to the program (e.g., adding dead code, adding/removing statements, etc.).

%

\subsubsection{Reducing the Search Space}
The third technique is orthogonal to the first two and aims to reduce the search space of relevant modifications a priori, rather than searching it efficiently.
Concretely, for a position $l_i$ in the program $p$ at which the prediction is made, it refines the set of valid program modifications as $\Delta(x) \subseteq \prod_{l_j} \Delta(p, l_j)$ for all positions $\{l_j \mid l_j \in l_{1:n} \wedge \scode{reachable}(l_j, l_i) \}$.
Here, we use $\scode{reachable}(l_j, l_i)$ to denote that position $l_j$ can affect position $l_i$. When representing programs as graphs, this can be computed a priori by checking the reachability between the two corresponding nodes.
Additionally, when used together with gradient based optimization, such check is not necessary as the gradients will naturally be zero.
To obtain a program representation where dependencies between many program locations are removed, we learn to refine program representation as described in Section 3 and Appendix D.

\paragraph{Limitations and advantages}
The main advantage of this approach is that it applies to both renaming and structural modifications.
The main disadvantage is that it depends on the fact the dependencies between program locations can be check efficiently and learned as part of the training. While we show how this can be done for graph neural networks, our approach currently does not support other models such as recurrent neural networks.

\paragraph{Summary}
In this section, we described how adversarial attacks can be applied to code via set of program modifications.
The adversarial attacks we consider are applied on the discrete input (i.e., the attack always correspond to a concrete program) rather than considering attacks in the latent space that are not directly interpretable.
We describe two existing techniques that can be used to guide the search for adversarial attacks (greedy search and gradient-based search) and one makes the attacks easier by reducing the search space. 
As such, these techniques are quite general and can be applied to number of tasks over code.
In our experiments, we use the greedy search technique together with reducing the search space.

\section{Learning to Refine Representation}\label{ilp}
In this section, we provide formal definition of the integer linear program (ILP) encoding used to solve the optimization problem presented in Section 3. 
Recall, that the problem statement is as follows.

\begin{figure*}[t]
\centering
\begin{tikzpicture}
\node (obj) { $\text{minimize} \quad \sum_{q \in \Phi} cost_{q} \quad \text{subj. to}$ };

\node[below=of obj, xshift=-1.3cm, yshift=1.2cm] {\footnotesize $\forall (\langle V, E, \xi_V, \xi_E \rangle, y)\!\in\!\dataset$};

\node (constraints) [right=of obj, xshift=-1.2cm] {
\begin{tabular}{l l l}               
$\quad 0 \leq f_{st} \leq cost_{\phi(\langle s, t\rangle)}$ & $\forall \langle s, t \rangle \in E$ & \footnotesize [edge capacity]\\
$\quad r_v + \sum_{\{s \mid (s, v) \in E\}} f_{sv} = \sum_{\{ t \mid (v, t) \in E\}} f_{vt}$ &  $\forall v \in V$ & \footnotesize [flow conservation]
\end{tabular}
};
        
\end{tikzpicture}
\vspace{-0.7em}
\caption{Formulation of the refinement problem from \Eqref{refinement_statement} as a minimum cost maximum flow integer linear program.}
\figlabel{ilp}
\end{figure*}

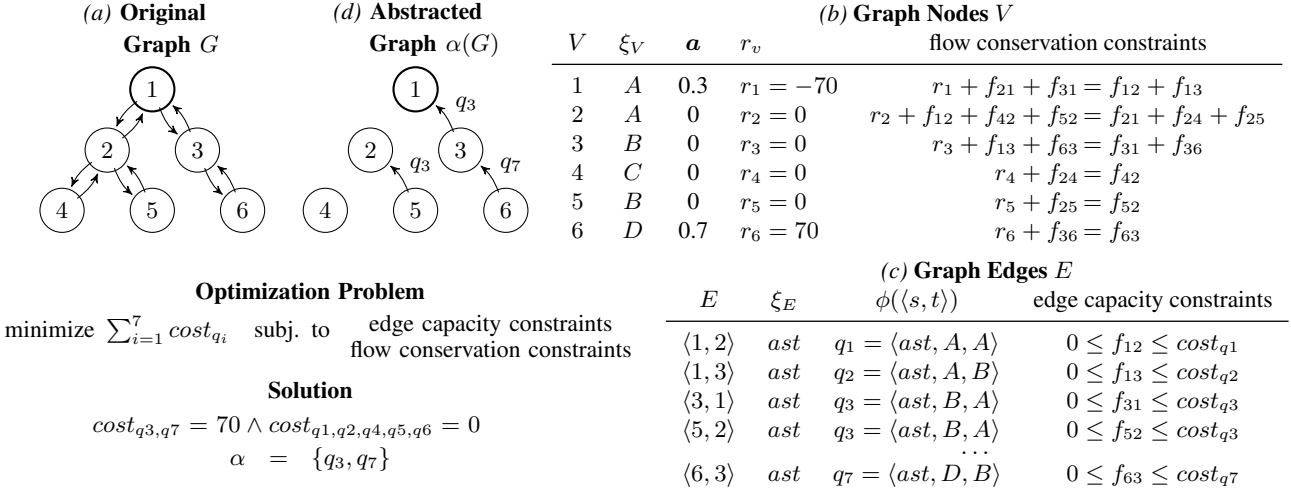
\begin{figure*}[t]
\centering
\begin{tikzpicture}[trim left=-2.05cm,>=stealth',bend angle=15,auto]

\tikzstyle{place}=[circle,thin,draw=black,minimum size=5mm, font=\footnotesize]
\tikzstyle{con}=[circle,thin,minimum size=5mm,font=\footnotesize]
\tikzstyle{label}=[font=\footnotesize]

\begin{scope}

\node [place, thick] (a) {$1$};
\node [place, below=of a, yshift=0.8cm, xshift=-0.6cm] (b) {$2$};
\node [place, below=of a, yshift=0.8cm, xshift=0.6cm] (c) {$3$};
\node [place, below=of b, yshift=0.8cm, xshift=-0.6cm] (d) {$4$};
\node [place, below=of b, yshift=0.8cm, xshift=0.6cm] (e) {$5$};	
\node [place, below=of c, yshift=0.8cm, xshift=0.6cm] (f) {$6$};

    
\node [con] at (a) {}
	edge [post,bend right] (b)
	edge [post,bend right] (c);
	
\node [con] at (b) {}
	edge [post,bend right] (a)
	edge [post,bend right] (d)
	edge [post,bend right] (e);
	
\node [con] at (c) {}
	edge [post,bend right] (a)
	edge [post,bend right] (f);

\node [con] at (d) {}
	edge [post,bend right] (b);
	
\node [con] at (e) {}
	edge [post,bend right] (b);
	
\node [con] at (f) {}
	edge [post,bend right] (c);
	
\node[above=of a, xshift=-0.1cm, yshift=-1.0cm, text width=5cm, align=center] { \footnotesize \captionn{a} \bf Original \\ ~~~~~~Graph $G$};
\end{scope}

\node (nodes) [right=of a, xshift=3.8cm, yshift=-0.46cm, font=\small] {
\tra{1.1}
\begin{tabular}{c c c l r@{\hskip 0.02in} l}    
\multicolumn{6}{c}{\footnotesize \captionn{b} \bf Graph Nodes $V$}\\
$V$ & $\xi_V$ & $\va$ & $r_v$ & \multicolumn{2}{c}{flow conservation constraints} \\ \midrule 
$1$ & $A$ & 0.3 & $r_1 = -70$  & $r_1 + f_{21} + f_{31}$&$= f_{12} + f_{13}$ \\
$2$ & $A$ & 0 & $r_2 = 0$ & $r_2 + f_{12} + f_{42} + f_{52}$ & $= f_{21} + f_{24} + f_{25}$  \\
$3$ & $B$ & 0 & $r_3 = 0$ & $r_3 + f_{13} + f_{63}$ & $= f_{31} + f_{36}$  \\
$4$ & $C$ & 0 & $r_4 = 0$ & $r_4 + f_{24}$ & $= f_{42}$ \\
$5$ & $B$ & 0 & $r_5 = 0$ & $r_5 + f_{25}$ & $= f_{52}$ \\
$6$ & $D$ & 0.7 & $r_6 = 70$ & $r_6 + f_{36}$ & $= f_{63}$ \\
\end{tabular}
};


\node (edges) [below=of nodes, yshift=1.1cm, xshift=0.8cm, font=\small] {
\tra{1.1}
\begin{tabular}{c c c c}
\multicolumn{4}{c}{\footnotesize \captionn{c}  \bf Graph Edges $E$}\\
$E$ & $\xi_E$ & $\phi(\langle s, t \rangle)$ &  edge capacity constraints  \\ \midrule 
$\langle 1, 2\rangle$ & $ast$ & $q_1 = \langle ast, A, A \rangle$ & $0 \leq f_{12} \leq cost_{q1}$ \\
$\langle 1, 3\rangle$ & $ast$ & $q_2 = \langle ast, A, B \rangle$ & $0 \leq f_{13} \leq cost_{q2}$ \\
$\langle 3, 1\rangle$ & $ast$ & $q_3 = \langle ast, B, A \rangle$ & $0 \leq f_{31} \leq cost_{q3}$ \\
$\langle 5, 2\rangle$ & $ast$ & $q_3 = \langle ast, B, A \rangle$ & $0 \leq f_{52} \leq cost_{q3}$ \\[-0.5em]
\multicolumn{4}{c}{$\dots$}  \\[-0.1em]
$\langle 6, 3\rangle$ & $ast$ & $q_7 = \langle ast, D, B \rangle$ & $0 \leq f_{63} \leq cost_{q7}$ \\
\end{tabular}
};

\node[left=of edges, yshift=0.8cm, xshift=0.7cm, text width=8.2cm, align=center, font=\small] (ilp) { {\footnotesize \bf Optimization Problem} \\[0.5em]
$\text{minimize} \; \sum_{i = 1}^{7} cost_{q_i} \quad \text{subj. to}\qquad \qquad \qquad \qquad \qquad \qquad $};

\node[below=of ilp, yshift=1.65cm, xshift=2.4cm, text width=8.2cm, align=center, font=\small]  { 
$\text{edge capacity constraints}$\\
$\text{flow conservation constraints}$
};

\node[below=of ilp, yshift=0.75cm, text width=6cm, align=center, font=\small] { {\footnotesize \bf Solution} \\[0.3em]
$cost_{q3, q7} = 70 \wedge cost_{q1, q2, q4, q5, q6} =0 \qquad $ \\[0.3em]
$\alpha\!=\!\{ q_3, q_7 \}$
};
%

\begin{scope}[xshift=3.5cm, yshift=0cm]

\node [place, thick] (a) {$1$};
\node [place, below=of a, yshift=0.8cm, xshift=-0.6cm] (b) {$2$};
\node [place, below=of a, yshift=0.8cm, xshift=0.6cm] (c) {$3$};
\node [place, below=of b, yshift=0.8cm, xshift=-0.6cm] (d) {$4$};
\node [place, below=of b, yshift=0.8cm, xshift=0.6cm] (e) {$5$};	
\node [place, below=of c, yshift=0.8cm, xshift=0.6cm] (f) {$6$};			
    
\node[font=\footnotesize, right=of a, xshift=-0.9cm, yshift=-0.2cm] {$q_3$};
\node[font=\footnotesize, right=of c, xshift=-0.9cm, yshift=-0.2cm] {$q_7$};
\node[font=\footnotesize, right=of b, xshift=-0.9cm, yshift=-0.2cm] {$q_3$};
	
	
\node [con] at (c) {}
	edge [post,bend right] (a);
	
\node [con] at (e) {}
	edge [post,bend right] (b);
	
\node [con] at (f) {}
	edge [post,bend right] (c);
	
\node[above=of a, yshift=-1.0cm, xshift=-0.1cm, text width=5cm, align=center] { \footnotesize \captionn{d} \bf Abstracted \\~~~~~ Graph $\alpha(G)$};
\end{scope}

\end{tikzpicture}
\vspace{-1.3em}
\caption{Illustration of ILP encoding from \Figref{ilp} on a~single graph where the prediction should be made for node~$1$.}
\figlabel{ilp_example}
\end{figure*}

\paragraph{Problem statement} Minimize the expected size of the refinement $\alpha \subseteq \Phi$ subject to the constraint that the expected loss of the model $f$ stays approximately the same:
\begin{equation}\label{refinement_statement}
\argmin_{\alpha \subseteq \Phi}  \sum_{(x, y) \in \dataset} \abs{\alpha(x)}
\end{equation}
\text{subject to}
\begin{equation*}
\textstyle \sum_{(x,y) \in \dataset} \ell(f(x), y) \approx \sum_{(x,y) \in \dataset} \ell(f(\alpha(x)), y)
\end{equation*}

Our problem statement is quite general and can be directly instantiated by: \captionn{i} using $\ell_{\scode{AbstainCrossEntropy}}$ as the loss (\Appref{uncertainty}), and \captionn{ii} using \textit{adversarial risk} (\Appref{adversarial}).

The motivation of solving \Eqref{refinement_statement} by phrasing it as ILP problem is that existing off-the-shelve ILP solvers can solve it efficiently and produce the optimal solution. 
We discuss an alternative end-to-end solution that does not depend on an external ILP solver at the end of Section 3.

\paragraph{Optimization via integer linear programming}
To solve Equation 2 efficiently, the key idea is that for each sample $(x, y)\!\in\!\dataset$ we: \captionn{i} capture the relevance of each node to the prediction made by the model~$f$ by computing the \textit{attribution} $\va(f, x, y)\!\in\!\sR^{\abs{V}}$ (as described in Section 3), 
and \captionn{ii} include the minimum number of edges necessary for a~path to exist between every relevant node (according to the attribution $\va$) and the  node where the prediction is~made. 
Preserving all paths between the prediction and relevant nodes encodes~the constraint that the expected loss stays approximately the~same.

Concretely, let us define a \textit{sink} to be the node for which the prediction is being made while \textit{sources} are defined to be all nodes $v$ with \textit{attribution} $a_v\!>\!t$.
Here, the threshold $t \in \sR$ is used as a form of regularization.
To encode the \textit{sources} and the \textit{sink} as an ILP program, we define an integer variable $r_v$ associated with each node $v \in V$ as:
\begin{equation*}
r_v =
\begin{cases}
   \;\; -\sum_{v' \in V \setminus\{v\}} r_{v'}	& \text{if } v~\text{is predicted node}  \text{~[\textit{sink}]}\\
   \;\; \floor{100 \cdot a_v}      	& \text{else if } a_v > t \qquad \;\text{~[\textit{sources}]}\\
   \;\; 0  							&  \text{otherwise} \\
\end{cases}
\end{equation*}

That is, $r_v$ for a source is its attribution value converted to an integer and $r_v$ for a sink is a negative sum of all source values.
Note that in our definition it is not possible for a~single node to be both \textit{source} and a \textit{sink}.
For cases when the \textit{sink} node has a non-zero attribution, this attribution is simply left out since every node is trivially connected to~itself.

We then define our ILP formulation of the problem as shown in \Figref{ilp}.
Here $cost_q$ is an integer variable associated with each edge feature and denotes the edge capacity (i.e., the maximum amount of flow allowed to go trough the edge with this feature), $f_{st}$ is an integer variable denoting the amount~of~flow over the edge $\langle s, t\rangle$, the constraint $0\!\leq\!f_{st}\!\leq cost_{\phi(\langle s, t\rangle)}$ encodes the edge capacity, and $r_v + \sum_{\{s \mid (s, v) \in E\}} f_{sv} = \sum_{\{ t \mid (v, t) \in E\}} f_{vt}$ encodes the flow conservation constraint which requires that the flow generated by the node $r_v$ together with the flow from all the incoming edges $\sum_{\{s \mid (s, v) \in E\}} f_{sv}$ has to be the same as the flow leaving the node $\sum_{\{ t \mid (v, t) \in E\}} f_{vt}$.
The solution to this ILP program is a $cost$ associated with each edge feature $q\in\Phi$.
If the cost for a given edge feature is zero, it means that this feature was not relevant and can be removed.
As a~result, we define the refinement $\alpha = \{q \mid q \in \Phi \wedge cost_q > 0 \}$ to contain all edge features with non-zero weight.

\paragraph{Example}
As a concrete example, consider the initial graph shown in \Figref{ilp_example}\emph{a} and assume that the prediction is made for node $1$.
For simplicity, each node has a~single attribute~$\xi_V$, as shown in \Figref{ilp_example}\emph{b}, and all edges are of type $ast$.
The edge feature for edge $\langle 1, 3 \rangle$ is therefore $\langle ast, A, B \rangle$, since $\xi_E(\langle 1, 3 \rangle)\!=\!ast$, $\xi_V(1)\!=\!A$ and $\xi_V(3)\!=\!B$,  as shown in \Figref{ilp_example}\emph{c}.
The \textit{attribution} $\va$ reveals two relevant nodes for this prediction -- the node itself with score $0.3$ and node $6$ with score $0.7$.
We therefore define a~single source $r_6\!=\!70$ and a~sink $r_1\!=\!-70$ and encode both the edge capacity constraints, and the flow conservation constraints as shown in \Figref{ilp_example} (note that according to \Figref{ilp}, we would encode all samples in $\dataset$ jointly).
The minimal cost solution assigns cost $70$ to edge features $q_3$ and $q_7$ which are needed to propagate the flow from node $6$ to node $1$.
The graph obtained by applying the abstraction $\alpha\!=\!\{q_3, q_7\}$ is shown in \Figref{ilp_example}\emph{d} and makes the prediction independent of the subtree rooted at node~$2$.
Notice however, that an additional edge is included between nodes $5$ and $2$. This is because $\alpha$ is computed using \textit{local} edge features~$\phi$ only, which are the same for edges $\langle 3, 1 \rangle$~and~$\langle 5, 2 \rangle$.

%
%
%
%
%

\end{document}